\documentclass[preprint,11pt,3p,authoryear]{elsarticle}
\usepackage{nomencl}
\makenomenclature
\usepackage[draft]{changes} % Use 'draft' to show changes and 'final' to show final version
\definechangesauthor[name={default}, color=purple]{default}

\usepackage{natbib}
\usepackage{float}
\usepackage{svg}
\usepackage{url}
\usepackage{amssymb}
\usepackage{amsmath}
\usepackage{gensymb}
\usepackage{booktabs}
\usepackage{multirow}
\usepackage{threeparttable}
\usepackage{lineno}
\usepackage{graphicx}
\usepackage{caption}
\usepackage{array}
\usepackage{subcaption}
\usepackage{cleveref}
\usepackage{amsthm}
\usepackage{geometry}
\usepackage{tabularray}
\usepackage{setspace}
\usepackage{longtable}
\begin{document}
\doublespacing
\begin{frontmatter}

% \linenumbers

\title{
Operator Learning for Consolidation: An Architectural Comparison for DeepONet Variants
}

\author[label1,label2]{Yongjin Choi}
\author[label1]{Chenying Liu}
\author[label1]{Jorge Macedo}

\affiliation[label1]{
    organization={School of Civil and Environmental Engineering, Atlanta, Georgia Institute of Technology}, 
    city={Atlanta}, 
    postcode={30332}, 
    state={GA}, 
    country={USA}
}
\affiliation[label2]{
    organization={InnoCORE PRISM-AI Center, Korea Advanced Institute of Science and Technology (KAIST)}, 
    city={Daejeon}, 
    postcode={34141},  
    country={Republic of Korea}
}

% \author{Anonymous Authors}
% \affiliation[label1]{
%     organization={Anonymous Institution}
% }

\begin{abstract}

Deep Operator Networks (DeepONets) have emerged as a powerful surrogate modeling framework for learning solution operators in PDE-governed systems. While their use is expanding across engineering disciplines, applications in geotechnical engineering remain limited. This study systematically evaluates several DeepONet architectures for the consolidation problem. We initially consider three architectures: a standard DeepONet with the coefficient of consolidation embedded in the branch net (Models 1 and 2), and a physics-inspired architecture with the coefficient embedded in the trunk net (Model 3). Results show that Model 3 outperforms the standard configurations (Models 1 and 2) but still has limitations when the target solution (excess pore pressures) exhibits significant variation. To overcome this limitation, we propose a Trunknet Fourier feature-enhanced DeepONet (Model 4) that addresses the identified limitations by capturing rapidly varying functions. We further extend Model 4 to 3D scenarios. Although the computational speedup can be modest in the 1D case (1.5-100× compared with traditional solvers), the speedup becomes more pronounced in 3D, reaching approximately 1,000×. Leveraging this efficiency, we offer a conceptual demonstration of DeepONet’s potential to accelerate uncertainty quantification in a 3D consolidation problem. Overall, the study highlights the potential of DeepONets to enable efficient, generalizable surrogate modeling in geotechnical applications, advancing the integration of scientific machine learning in geotechnics, which is at an early stage.

\end{abstract}

\begin{keyword}
Scientific machine learning \sep
Neural Operator \sep
DeepONet \sep 
Terzaghi's Consolidation PDE \sep
Surrogate Modeling \sep
Fourier Feature Embedding \sep

\end{keyword}

\end{frontmatter}

% \linenumbers
\section{Introduction}

\label{sec:intro}

The modeling and simulation of geotechnical problems typically rely on solving partial differential equations (PDEs) or ordinary differential equations (ODEs) using numerical methods such as the Finite Element Method (FEM) or the Finite Difference Method (FDM). While these approaches are well-established, they often incur significant computational costs.

Recent advances in scientific machine learning \citep{faroughi2022sciml} have introduced computationally efficient deep learning-based surrogate modeling frameworks \citep{huang2025deeplearningpdes} to approximate PDE solutions obtained via numerical solvers. Physics-Informed Neural Networks (PINNs) \citep{raissi2019pinn, cuomo2022sciml_pinn} constitute one of these emerging frameworks. PINNs approximate PDE solutions by enforcing governing equations as soft constraints within the loss function during neural network training. Although PINNs have been widely explored in various engineering domains, their application in geotechnical engineering remains relatively limited, with recent studies focusing on assessing the potential of PINN. For example, \cite{bekele2021pinn_consolidation_jrmge} evaluated the PINN potential for forward and inverse problems in soil consolidation. \cite{gao2023pinn_seepage_phreatic_line, yang2025seepage_swcc} applied PINN to seepage analysis and soil-water characteristic curves. \cite{liu2025leveraging} assessed PINN's potential in seismic site response analysis, leveraging Fourier feature embedding. 

Although PINNs have shown promise for solving PDEs across many engineering disciplines, recent studies \citep{grossmann2024pinn_vs_fem, bueno2025pinn_vs_fdm} indicate that they are not necessarily more effective PDE solvers than traditional numerical methods. PINNs are designed to approximate solutions for a fixed set of PDE coefficients and boundary conditions, which means that they require retraining if those conditions change \citep{faroughi2022sciml}. This retraining process is computationally expensive \citep{sharma2022accelerated_pinn}, posing challenges for using PINNs in applications with variable system configurations. Additionally, PINNs require an explicit formulation of governing equations, which limits their applicability in domains where physical behavior is described by empirical or phenomenological models rather than well-defined PDEs. 

Neural operators \citep{kovachki2023neural_operators, li2020fourierNOs, lu2021learning, li2020graphNOs} have emerged as an alternative in scientific machine learning that address some of PINNs' limitations. Neural operators are a new type of deep learning framework designed to learn operators that map between infinite-dimensional function spaces. Unlike conventional deep learning methods that approximate solutions for specific inputs, neural operators learn the underlying solution operator that maps input functions (e.g., material property fields, boundary conditions, loadings) to output solution fields. This operator-based learning enables neural operators to generalize across varying initial and boundary conditions without retraining. Additionally, neural operators do not require explicit knowledge of the governing equations, making them particularly well-suited to complex physical phenomena in which a closed-form PDE is not defined.  
Compared to conventional numerical methods or PINNs, which need to solve the underlying PDEs for each new set of conditions \citep{koric2024deeponet_plasticity, jiang2024mionet_carbon, faroughi2022sciml}, neural operators offer significantly more efficient inference based on learned function mapping. These advantages make them particularly well-suited to serve as surrogate models in engineering applications with complex operating conditions.

Among various neural operator architectures, Deep Operator Networks (DeepONet), proposed by \cite{lu2021learning}, have gained substantial attention due to their formulation. DeepONet is based on the universal approximation theorem for operators \citep{chen1995universal}, which establishes its ability to approximate nonlinear solution operators of PDEs given sufficient network complexity and data coverage. This framework allows DeepONet to efficiently learn generalizable operators \citep{lu2022comprehensive}.

%Compared to other deep learning architectures applied to systems governed by PDEs, 
DeepONet can achieve greater data efficiency, faster training convergence, and improved accuracy \citep{lu2021learning} compared to conventional architectures, such as feed-forward networks (FNNs), sequence-to-sequence models (seq2seq), and residual networks (ResNets) in operator learning tasks. Furthermore, the architecture of DeepONet offers interpretability through its explicit separation of functional roles: the branch net learns to encode the input function, representing the coefficients of a latent basis, while the trunk net learns to output corresponding basis functions over the spatio-temporal domain, which are then combined to form a solution. This design mirrors the structure of functional mappings in infinite-dimensional spaces, where solution operators can be viewed as weighted combinations of basis functions parameterized by input functions \citep{naylor1982linear_operator_theory}. This formulation aligns with the mathematical framework of operator learning \citep{kovachki2023neural_operators}.

DeepONet has been applied in various fields, such as fluid mechanics \citep{kumar2024deeponet_engine, bai2024deeponet_cfd}, material science \citep{yin2022deeponet_solid}, structural mechanics \citep{he2023deeponet_structure, ahmed2024deeponet_structure}, and thermodynamics \citep{koric2023deepoent_heat, kushwaha2024deeponet_heat}. These efforts have highlighted DeepONet's potential in modeling complex physical processes. However, the potential of DeepONet in geotechnical engineering remains largely unexplored. This study aims to introduce DeepONet as a surrogate modeling approach for geotechnical problems, focusing on the consolidation phenomenon, one of the most prevalent geotechnical engineering problems. The motivation for exploring the consolidation problem is that it can serve as a benchmark to demonstrate DeepONet's capabilities to the geotechnical community, while also highlighting the role of the different architectural adaptations explored in this study.

More specifically, this study presents a systematic investigation into DeepONet architectural choices tailored for the consolidation phenomenon. We first explain the standard DeepONet architecture, then explore three learning architectures to gain initial insights. The first two models follow DeepONet variants previously investigated \citep{tan2022enhanced_deeponet,jin2022mionet}, and the third model adopts a non-traditional architectural choice, with its improved accuracy supported by the analytical solution structure of the consolidation problem. Insights and challenges identified across the explored architectures are shared and used to propose a DeepONet architecture enhanced with TrunkNet Fourier feature embedding, which demonstrates superior performance. We further test the scalability in 3D consolidation scenarios, and present a conceptual application to uncertainty quantification. Overall, this study contributes to the integration of advanced scientific machine learning into geotechnical applications.

\section{Deep operator networks (DeepONet)}

DeepONet approximates mappings between infinite-dimensional function spaces. This capability positions DeepONet as a powerful framework for operator learning. Formally, consider an operator $\mathcal{G}$ \nomenclature{$\mathcal{G}$}{The solution operator that maps input functions to output functions} defined as:

\begin{equation}
    \mathcal{G}:\mathcal{U}\ni {u}\mapsto \mathcal{S}\ni s
    \label{eq:operator}
\end{equation}

where $\mathcal{U}$ \nomenclature{$\mathcal{U}$}{The infinite-dimensional input function space} and $\mathcal{S}$ \nomenclature{$\mathcal{S}$}{The infinite-dimensional output function space} represent infinite-dimensional function spaces. Specifically, the operator $\mathcal{G}$ takes an input function $u\in \mathcal{U}$ \nomenclature{$u$}{The input function (element of $\mathcal{U}$)} and maps it to an output function $s\in \mathcal{S}$ \nomenclature{$s$}{The output function (element of $\mathcal{S}$)}. DeepONet aims to approximate $\mathcal{G}$ through a neural network model $\mathcal{G}_{\Theta}$ \nomenclature{$\mathcal{G}_{\Theta}$}{The neural network approximation (DeepONet) of the operator $\mathcal{G}$}, such that:

\begin{equation}
    \mathcal{G}_\Theta(u)(y) \approx s(y), \quad \text{for all } y \in D,
\label{eq:deeponet}
\end{equation}

Here, $y$ \nomenclature{$y$}{An evaluation location (point) in the output domain $D$} denotes a point within the spatial or temporal domain $D$ of interest. The input function $u$ is sampled at a fixed set of $m$ \nomenclature{$m$}{The number of fixed sensor locations for the input function} locations, ${\{x_\ell\}_{\ell=1}^m}$ \nomenclature{$x_\ell$}{The $\ell$-th fixed sensor location for sampling the input function $u$}, referred to as sensor locations. $\Theta$ \nomenclature{$\Theta$}{The set of learnable parameters in the neural network $\mathcal{G}_{\Theta}$} is a set of learnable parameters of the network.

\begin{figure}
    \centering
    \includegraphics[width=0.7\textwidth]{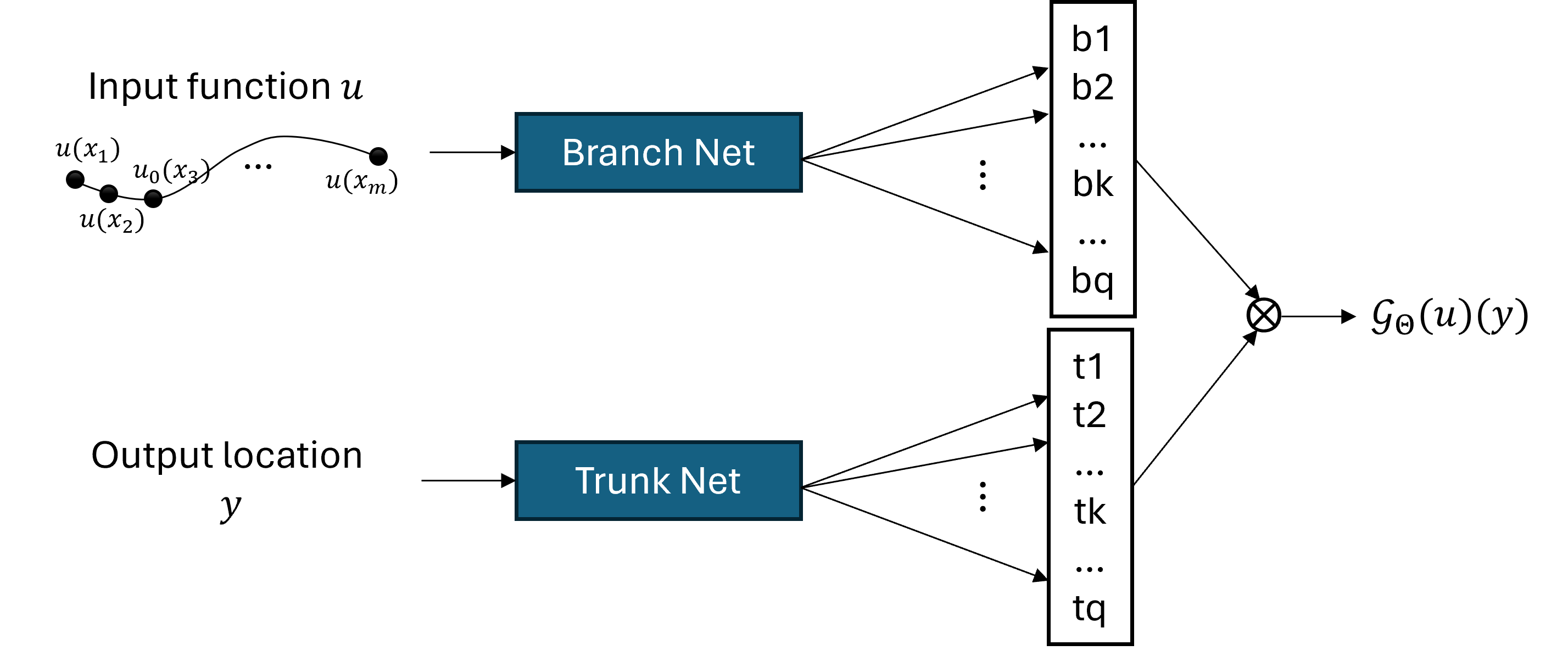}
    \caption{A standard DeepONet architecture.}
    \label{fig:generic-arch}
\end{figure}

Structurally, DeepONet comprises two subnetworks: the branch network (branch net) and the trunk network (trunk net) (see \Cref{fig:generic-arch}). The branch net encodes the input function $ u $, producing a latent feature representation $ \boldsymbol{b}(u) \in \mathbb{R}^{q} $ \nomenclature{$\boldsymbol{b}$}{The latent feature vector from the branch net output}. The trunk net encodes the evaluation location $ y $, generating another latent feature representation $ \boldsymbol{t}(y) \in \mathbb{R}^{q} $\nomenclature{$\boldsymbol{t}$}{The latent feature vector from the trunk net output}\nomenclature{$q$}{The output dimension of the branch and trunk nets (latent dimension) }. The final operator evaluation at a specific location $ y $ is obtained via the inner product of these two latent representations as follows:

\begin{equation}
    \mathcal{G}_\Theta(u)(y) = \sum_{k=1}^q b_k(u(x_1), u(x_2), ..., u(x_m)) t_k(y)
\label{eq:deeponet_architecture}
\end{equation}

where $ b_k(u) $ and $ t_k(y) $\nomenclature{$b_k$}{The $k$-th component of the branch net's output feature vector}\nomenclature{$t_k$}{The $k$-th component of the trunk net's output feature vector} denote the $k$-th component of the branch and trunk net outputs, respectively. The trunk and branch net are modeled with neural networks. The parameters of the neural networks comprise $\Theta$. The output of the trunk net can be interpreted as a basis for representing the output function, and the output of the branch net as the corresponding coefficients. DeepONet learns an optimal set of basis functions directly from data, allowing it to adapt to complex solution manifolds. 

DeepONet's learned basis and corresponding coefficients eliminate the need for manually designed basis functions, which are typically required in traditional numerical methods. For example, the finite element method uses local basis functions defined over a mesh, while spectral methods rely on global functions like sine and cosine series (e.g., Fourier series) chosen in advance \citep{williams2024deeponet_bases}. This flexibility allows DeepONet to effectively capture complex, nonlinear operator mappings in high-dimensional PDEs.

\section{Consolidation and associated PDEs}

Consolidation describes how excess pore water pressure (PWP) dissipates from saturated soils under load over time, causing gradual ground settlement, particularly in clayey soils. The governing PDE for 1D consolidation under vertical drainage (\Cref{eq:terzaghi_pde}) is derived from Darcy’s law for fluid flow through porous media and the effective stress principle.

\begin{equation}
    \frac{\partial u}{\partial t} = C_v \frac{\partial^2 u}{\partial z^2}
    \label{eq:terzaghi_pde}
\end{equation}

where $ u(z,t) $\nomenclature{$u(z,t)$}{The excess pore water pressure (PWP) at depth $z$ and time $t$} is the excess PWP at depth $ z $ and time $ t $, and $ C_v $\nomenclature{$C_v$}{The coefficient of consolidation} is the coefficient of consolidation, a parameter that combines the effects of soil permeability and compressibility.

\Cref{eq:terzaghi_pde} represents a diffusion-type process: pore pressure gradients drive vertical flow through the soil matrix, and as water escapes, the excess pore pressure gradually diminishes. This dissipation results in an increase in effective stress and ultimately leads to ground settlement.

A common drainage condition to solve equation \Cref{eq:terzaghi_pde}, which is used in this study, considers that drainage occurs only through the top surface of the soil, while the bottom boundary is considered impermeable. This leads to a combination of boundary conditions: a Dirichlet condition at the top $u(0,t) = 0$, and a Neumann condition at the bottom $\frac{\partial u}{\partial z}\bigg|_{z=H} = 0$\nomenclature{$H$}{The depth of soil}, which enforces a zero hydraulic gradient. The analytical solution to \Cref{eq:terzaghi_pde} is given by:

\begin{equation}
    u(z,t) = \sum_{a=0}^{\infty} \left[ \frac{2u_0}{A} \sin\left(\frac{Az}{H_{\mathrm{dr}}}\right) \right] e^{-A^2 T_v}
\label{eq:terzaghi_analytical_solution}
\end{equation}

Where $A = \frac{\pi}{2}(2a+1)$\nomenclature{$A$}{A term in the analytical solution: $A = \frac{\pi}{2}(2a+1)$} under single drainage boundary condition, $u_0$\nomenclature{$u_0$}{The initial excess PWP profile, $u(\cdot,0)$ (used as the input function $u$)} is the initial excess pore pressure, $H_{dr}$\nomenclature{$H_{dr}$}{The drainage path length} is the drainage path length, and $T_v=\frac{C_v t}{H_{dr}^2}$\nomenclature{$T_v$}{The dimensionless time factor: $T_v = C_v t / H_{dr}^2$} is the dimensionless time factor. This analytical form reveals that the solution is composed of spatial sine functions modulated by exponential decay terms dependent on $ C_v $ and time $ t $. This insight informs one of the DeepONet architecture choices that is later explored.

\section{Data generation and training}\label{sec:data_and_training}

We use the finite difference method to numerically solve \Cref{eq:terzaghi_pde}. The solution obtained is used to train the DeepONet and to test the performance. The finite difference method first discretizes the spatial domain uniformly into $ N_z $ nodes separated by a spacing $ \Delta z $. It then approximates the spatial second derivative at each internal node $ i $ ($ 1 \leq i \leq N_z - 2 $), using a second-order central finite-difference scheme (\Cref{eq:finite_diff}).

\begin{equation}
    \frac{\partial^2 u}{\partial z^2}\bigg|_{z=z_i} \approx \frac{u_{i+1} - 2 u_i + u_{i-1}}{\Delta z^2}.
    \label{eq:finite_diff}
\end{equation}

This discretization converts the PDE into a semi-discrete system of ODE (\Cref{eq:discretized_odes}).

\begin{equation}
    \frac{du_i}{dt} = C_v \frac{u_{i+1} - 2u_i + u_{i-1}}{\Delta z^2}, \quad 1 \leq i \leq N_z - 2.
    \label{eq:discretized_odes}
\end{equation}

For the single drainage boundary condition, the Dirichlet condition $u_0=0$ is applied at the top ($z=0$). At the impermeable bottom boundary ($z=H$, node $N_z-1$), the Neumann condition ($\frac{\partial u}{\partial z}|_{z=H}=0$) is applied by considering a ghost node $u_{N_z}$ and setting $u_{N_z}=u_{N_z-1}$. The ODE for the bottom boundary node $N_z- 1 $ thus becomes:

\begin{equation}
    \frac{du_{N_z-1}}{dt} = C_v\frac{u_{N_z-1}-2u_{N_z-1}+u_{N_z-2}}{\Delta z^2} = C_v\frac{u_{N_z-2}-u_{N_z-1}}{\Delta z^2}.
    \label{eq:bottom_node_single} % This label is retained from the original
\end{equation}

We employ both the implicit Backward Differentiation Formula (BDF) and the explicit Runge-Kutta (RK) methods to solve the discretized ODE system (see \Cref{eq:discretized_odes} and \Cref{eq:bottom_node_single}). BDF is used to generate data for DeepONet, while the RK method serves as a benchmark for comparative analysis of the computational performance of DeepONet and BDF. Both integrators are implemented using the \texttt{solve\_ivp} function from the SciPy library, with details included in \ref{sec:appendix}.

When generating data, we consider both uniform and spatially varying initial excess PWP profiles with depth $u(z, 0)$ as the initial condition. For the uniform profile, we randomly sample constant functions between 10 kPa and 20 kPa. For the spatially varying (non-uniform) profiles, we sample functions from a Gaussian Random Field (GRF), denoted as $u(z, 0) \sim G(\mu_r, C(z_1, z_2))$. The spatial correlation of this GRF is defined by a Gaussian covariance function:

\begin{equation}
    C(z_1, z_2) = \sigma_r^2 \exp\left(-\frac{||z_1 - z_2||^2}{l^2}\right)
\end{equation}

\nomenclature{$G(\mu_r, C)$}{Gaussian Random Field (GRF) with mean $\mu_r$ and covariance function $C(z_1, z_2)$}
\nomenclature{$C(\cdot,\cdot)$}{The covariance function of the Gaussian Random Field}
\nomenclature{$l$}{The correlation length (length-scale) of the GRF}
\nomenclature{$\mu_{r}$}{The mean of the Gaussian Random Field}
\nomenclature{$\sigma_r$}{The standard deviation of the Gaussian Random Field}

Here, $\mu_{r}$ is mean, $C(z_1, z_2)$ represents the covariance between two points at depths $z_1$ and $z_2$. The term $||z_1 - z_2||$ is the Euclidean distance between these points. The parameter $l$ is the correlation length (or length-scale), which controls the smoothness of the sampled functions; a larger $l$ generates smoother profiles. The term $\sigma_r$ is the standard deviation of the field, controlling the overall magnitude of the covariance. We use a $\sigma_r^2$ value of $1\,\text{kPa}^2$, a correlation length of $0.5$, and sample $\mu_r$ uniformly from the range $10$-$20\,\text{kPa}$. The coefficient of consolidation, $C_v$, is also sampled uniformly from the range $0.3$-$1.0\,\text{m}^2/\text{year}$. The spatial domain spans the normalized depth $\frac{z}{H_{dr}} \in [0, 1]$ and normalized time $T_v=\frac{C_v t}{H_{dr}^2} \in [0, 2]$.

In the DeepONet training, the model parameter set $\Theta$ in $\mathcal{G}_\Theta$ is optimized by minimizing the mean squared error between the predicted solution and the true solution at $y$ as shown in \Cref{eq:deeponet_loss}.

\begin{equation}
\begin{aligned}
    \mathcal{L}(\Theta) &= \frac{1}{NP} \sum_{i=1}^{N} \sum_{j=1}^{P} \left| \mathcal{G}_\Theta(u^{(i)})(y_j^{(i)}) - \mathcal{G}(u^{(i)})(y_j^{(i)}) \right|^2 \\
    &= \frac{1}{NP} \sum_{i=1}^{N} \sum_{j=1}^{P} \left| \sum_{k=1}^{q} b_k(u^{(i)}(x_1), \dots, u^{(i)}(x_m)) t_k(y_j^{(i)}) - \mathcal{G}(u^{(i)})(y_j^{(i)}) \right|^2
\end{aligned}\label{eq:deeponet_loss}
\end{equation}

\nomenclature{$N$}{The number of input functions (samples) in the DeepONet}
\nomenclature{$P$}{The number of evaluation locations (output points) per input function in the DeepONet}

where $\{u^{(i)}\}_{i=1}^N$ denotes $N$ separate input functions sampled from $\mathcal{U}$. For each $u^{(i)}$, $\{y_j^{(i)}\}_{j=1}^P$ are $P$ locations in the domain of $\mathcal{G}(u^{(i)})$, and $\mathcal{G}(u^{(i)})(y_j^{(i)})$ is the corresponding output data evaluated at $y_j^{(i)}$. Contrary to the fixed sensor locations of $\{x_{\ell}\}_{\ell=1}^m$, the locations of $\{y_j^{(i)}\}_{j=1}^P$ may vary for different $i$, thus allowing to construct a flexible and continuous representation of the output functions $s \in \mathcal{S}$.

\begin{figure}[!htbp]
    \centering
    \includegraphics[width=0.8\textwidth]{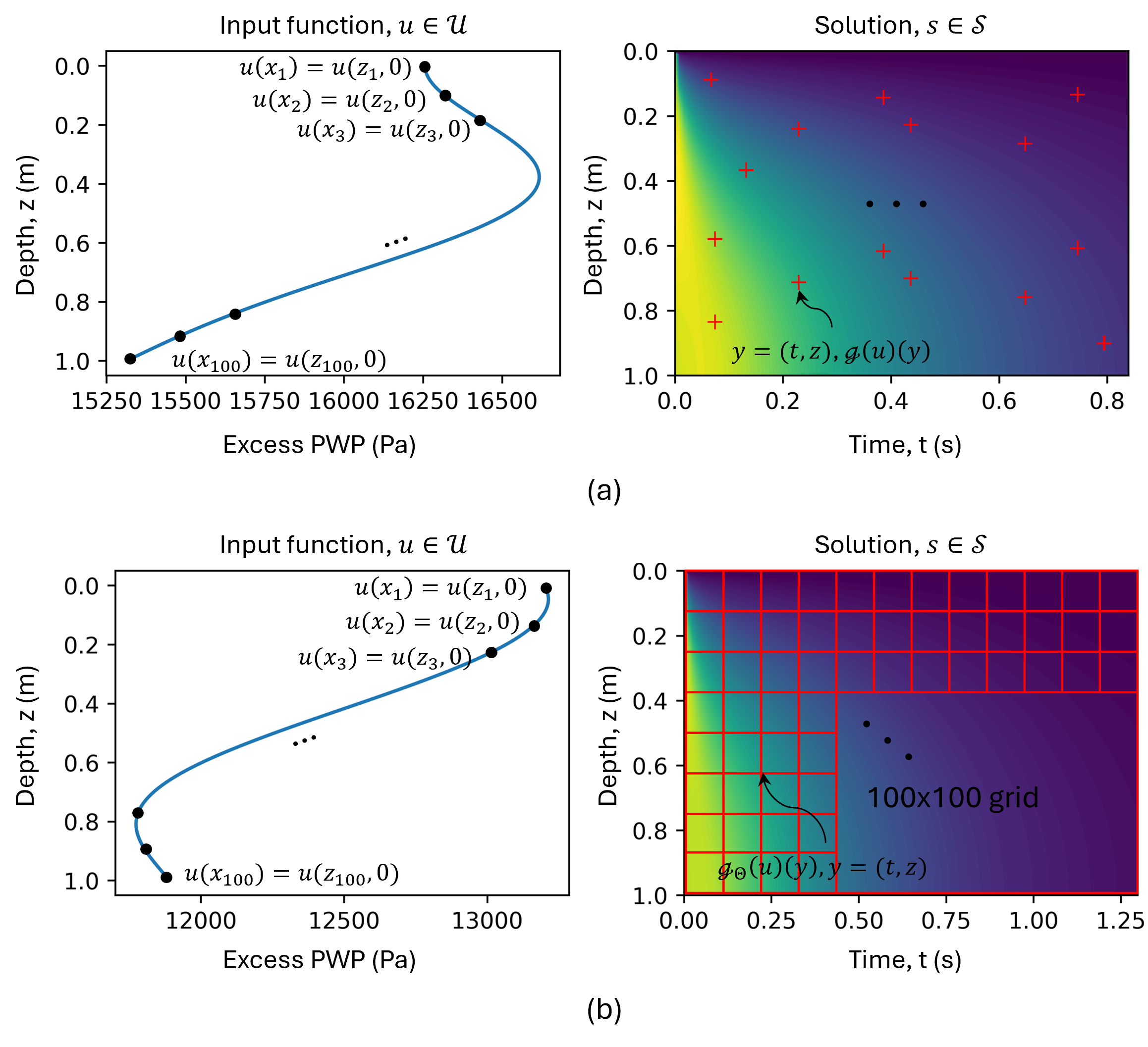}
    \caption{\textbf{(a) Training and Validation Data Sampling Method.} Input functions (initial excess pore water pressure (PWP)) are sampled at $m=100$ fixed equally spaced locations, denoted by $\{x_\ell \}_{\ell=1}^{100}$ (see left side figure). For each given input function $u^{(i)}$, the corresponding solution is sampled at $P=100$ randomly chosen output evaluation points, denoted by $\{y_j^{(i)}\}_{j=1}^{100}$ (the red markers on the right side figure). This results in 100 training examples of the form $(u^{(i)}, y_j^{(i)}, \mathcal{G}(u^{(i)})(y_j^{(i)}))$ for each input function. For the training data, we generate $N=40,000$ input functions, forming a total of $40,000 \times 100$ training examples. For validation, we generate $N=5,000$ input functions, forming $5,000 \times 100$ validation examples. \textbf{(b) Test Method.} Input functions unseen during training are sampled at $m=100$ fixed equally spaced locations, $\{x_\ell \}_{\ell=1}^{100}$ (see left side figure). For each given input function $u^{(i)}$, we evaluate the learned solution operator $\mathcal{G}_{\Theta}(u^{(i)})(y)$ at a dense $100 \times 100$ uniformly spaced grid of $y$ points across the output domain (see right side figure).}
    \label{fig:data}
\end{figure}

\added{We train and validate DeepONet using N=40,000 and N=5,000 functions and coefficients, respectively. The trained model is tested on 500 unseen functions and coefficients, with no overlap among the three sets.} Specifically, each input function represents an initial excess PWP and is discretized at m=100 fixed, equally spaced sensor locations, denoted by $ \{x_\ell\}_{\ell=1}^{100} $ (see the left side figure in \Cref{fig:data}a). That is, for each input function $ u^{(i)} $, the branch network receives the vector $ \left[u^{(i)}(x_1), u^{(i)}(x_2), \dots, u^{(i)}(x_{100})\right] $ as input. The corresponding output functions (i.e., solutions $ s = \mathcal{G}(u) $) are sampled at P=100 randomly selected evaluation points $ y = (z, t) $ (see the red markers in the right side figure in \Cref{fig:data}a), sampled independently for each input function. These locations are denoted $ \{y_j^{(i)}\}_{j=1}^{100} $, resulting in a total of $ 40,000 \times 100 $ training examples of the form $ \left(u^{(i)}, y_j^{(i)}, \mathcal{G}(u^{(i)})(y_j^{(i)})\right) $. The validation set is constructed in the same manner, with P=100 output evaluations per input, resulting in $ 5,000 \times 100 $ training examples. For testing, we assess the generalization capability of the learned operator $ \mathcal{G}_\Theta $ on 500 unseen input functions during training. Each is evaluated over a dense $ 100 \times 100 $ uniformly spaced grid in the output domain (see the right side figure in \Cref{fig:data}b), yielding P=10,000 output locations per function to visualize full solution fields. All input and output data are standardized using the mean and standard deviation computed from the training set.

\section{Model architectures}\label{sec:model_architecture}

\begin{figure}[!htbp]
    \centering
    \includegraphics[width=0.65\textwidth]{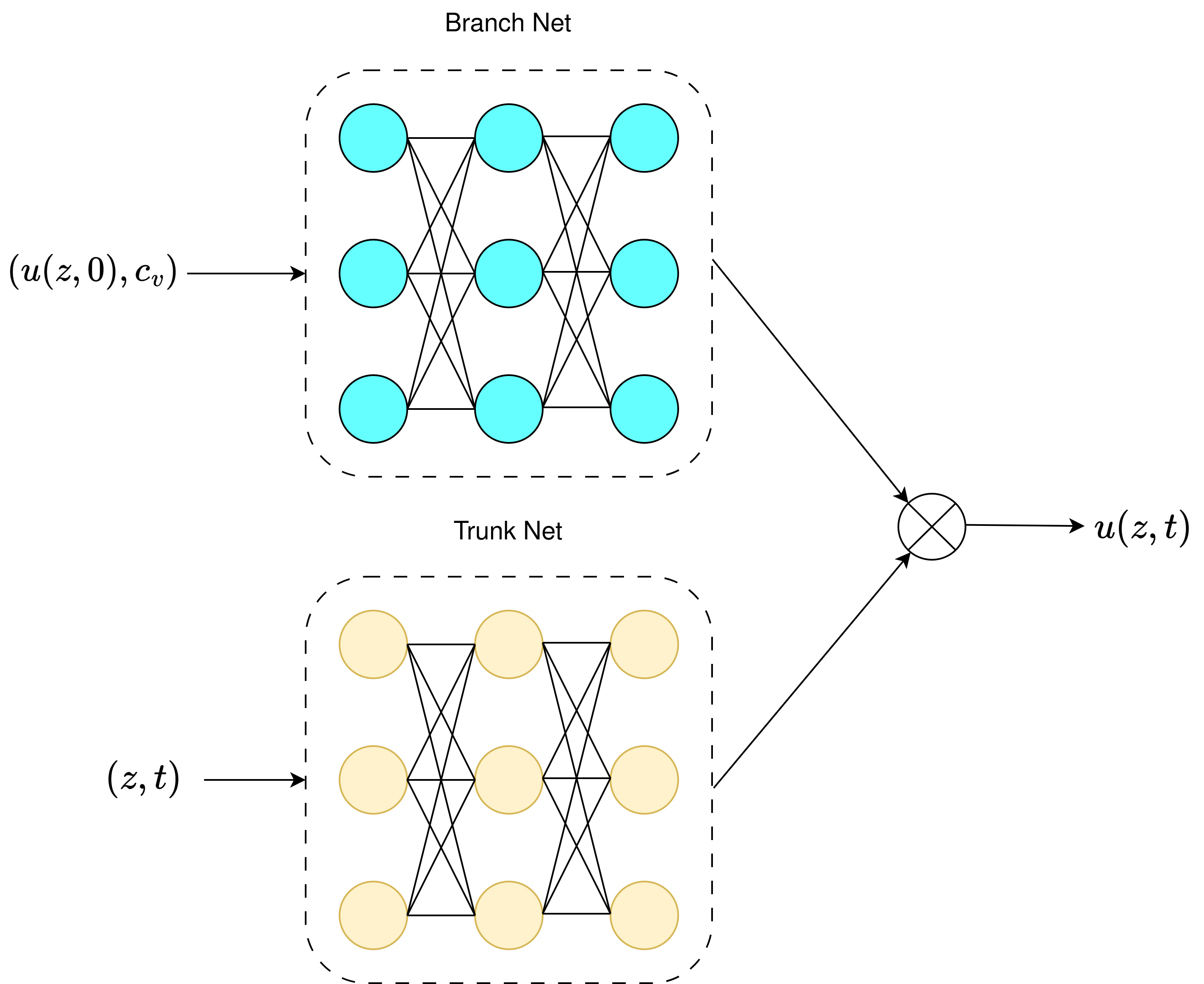}
    \caption{DeepONet with direct concatenation of $C_v$ to the branch net.}
    \label{fig:mod1}
\end{figure}

\begin{figure}[!htbp]
    \centering
    \includegraphics[width=0.65\textwidth]{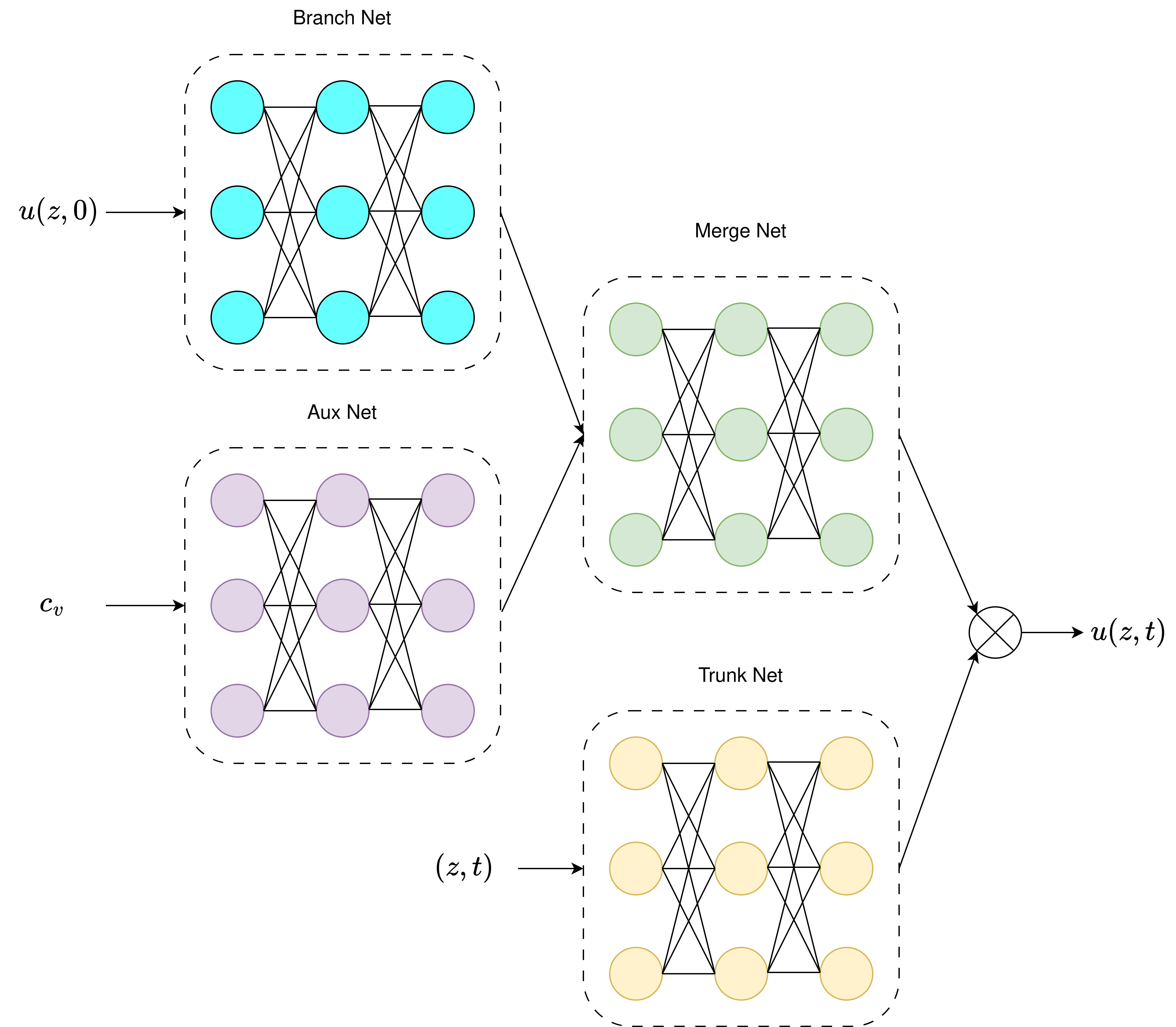}
    \caption{DeepONet with auxiliary networks for embedding $u(z,0)$ and $C_v$ and merging their representations.}
    \label{fig:mod2}
\end{figure}

\begin{figure}[!htbp]
    \centering
    \includegraphics[width=0.65\textwidth]{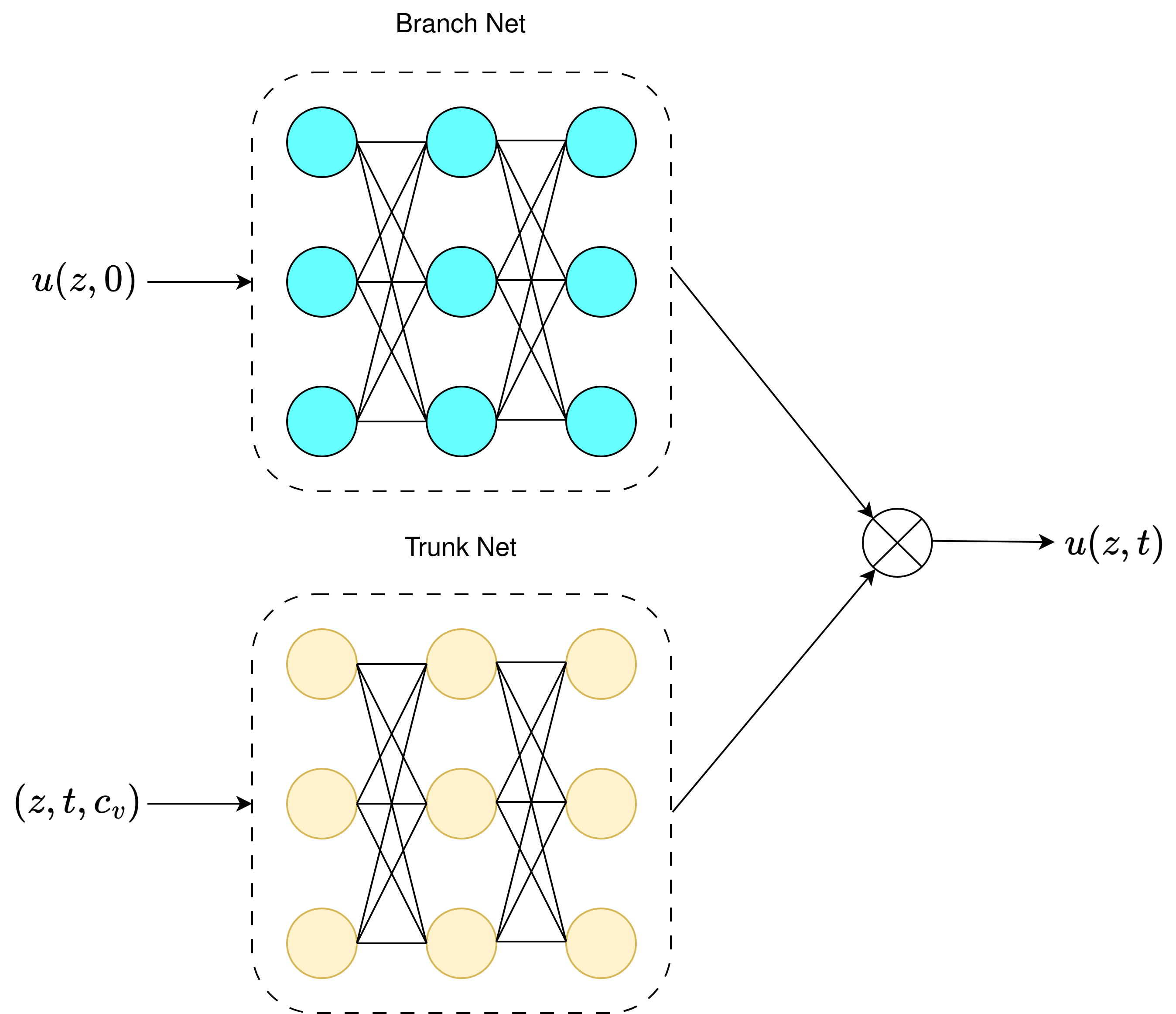}
    \caption{DeepONet with trunk net modulation by $C_v$.}
    \label{fig:mod3}
\end{figure}

We explore various model architectures, each reflecting a modeling assumption about the interaction between the PDE parameters, the input function, and the resulting solution. The motivation is to explore how the treatment and placement of governing parameters can affect performance. \Cref{fig:mod1,fig:mod2,fig:mod3} shows the different architectures investigated, and described as follows:

\begin{itemize}
    \item \textbf{Model 1 (\Cref{fig:mod1}):} This model serves as our baseline and adheres to the standard DeepONet formulation introduced by \cite{lu2021learning}. The branch net processes the input function, which in this case consists of the discretized initial PWP profile, $u(z,0)$. The scalar coefficient of consolidation, $C_v$, is concatenated with the discretized $u(z,0)$ vector, forming a single input tensor for the branch net. This design philosophy treats $C_v$ as a global conditioning parameter. The implicit assumption is that the branch net should learn a combined representation of how the initial PWP distribution and the material property $C_v$ jointly influence the consolidation process. The trunk net takes the spatiotemporal coordinates $(z,t)$ as input. 
    
    \item \textbf{Model 2 (Multiple-Input Operator Network) (\Cref{fig:mod2}):} This architecture modifies the branch net to handle $u(z,0)$ and $C_v$ separately before integrating their influence. One branch net is dedicated to encoding the initial PWP function, $u(z,0)$, while a second, auxiliary branch net processes the coefficient of consolidation, $C_v$, generating distinct latent representations for each. These two latent representations are then concatenated and passed through a "merge net" (another small neural network) to produce the final branch output vector. This approach allows the model to potentially learn more disentangled features for the initial condition and the PDE coefficient before their combined effect on the solution basis is determined. \cite{jiang2024mionet_carbon} reported a performance improvement by introducing the merge net. The trunk net architecture remains the same, taking $(z,t)$ as input. This model architecture aligns with the multiple-input operator network (MIONet) proposed by \cite{jin2022mionet}.
    
    \item \textbf{Model 3 (\Cref{fig:mod3}):} In Model 3, we propose feeding $C_v$ directly into the trunk net. The initial PWP function, $u(z,0)$, is processed by the branch net as before. However, the trunk net now receives an augmented input vector $(z, t, C_v)$. The rationale is that the trunk net learns to approximate the basis functions of the solution operator. By providing $C_v$ directly to the trunk, it can generate basis functions dynamically modulated by this physical parameter across the spatiotemporal domain. 
\end{itemize}

The performance of Models 1, 2, and 3 is discussed in \Cref{sec:results}, highlighting the challenges to predict excess PWP distribution during the early stages of consolidation, especially for non-uniform distributions across depth. This motivated the authors to explore an additional option. Specifically, an additional architecture option considers Model 3 enhanced with Fourier Feature Embedding \citep{tancik2020fourier}, which can help to learn rapidly varying functions (e.g., \cite{wang2021eigenvector_ffe}, \cite{liu2025leveraging}). 

\begin{figure}[!htbp]
    \centering
    \includegraphics[width=0.75\textwidth]{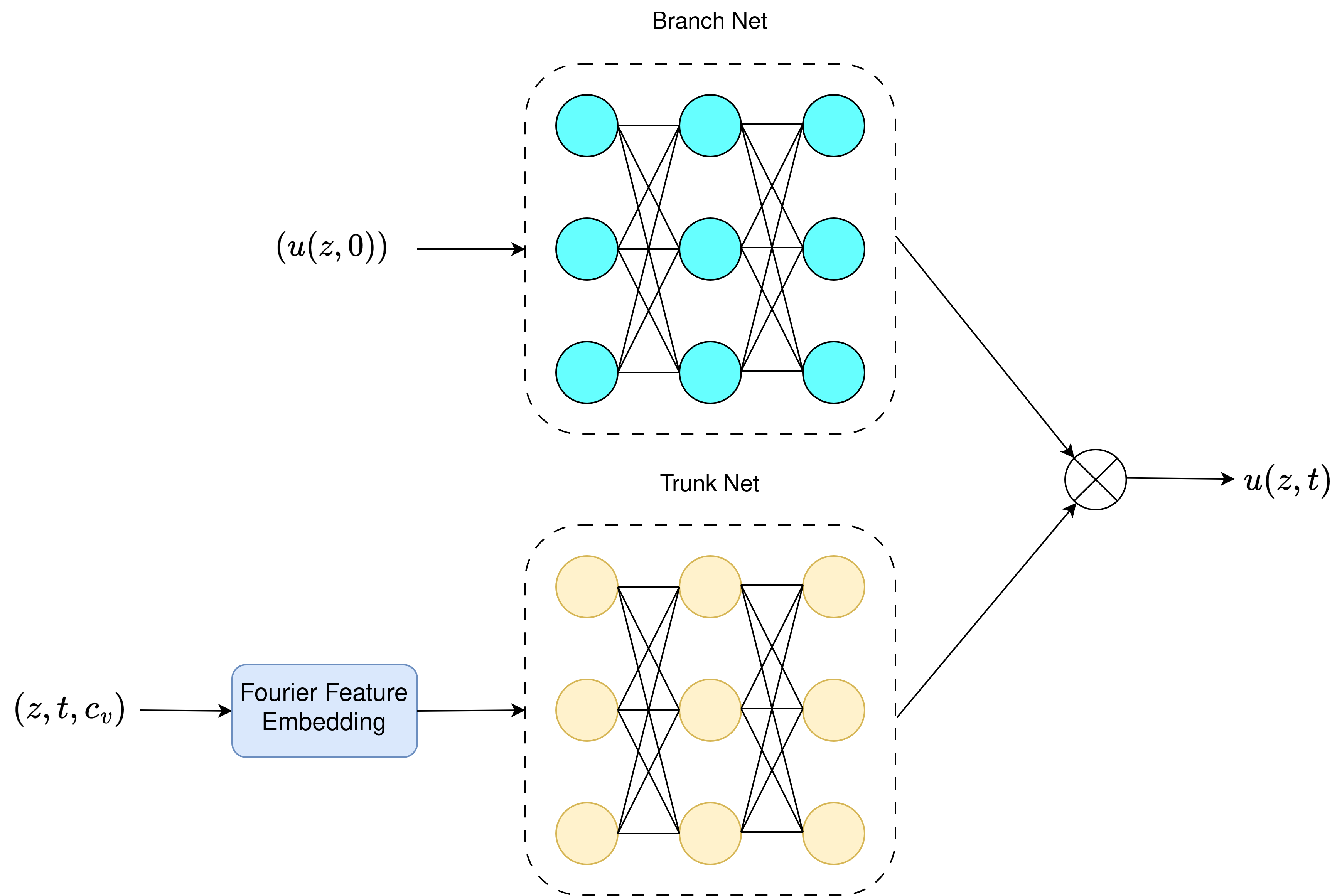}
    \caption{DeepONet with trunk net modulation by $C_v$ and Fourier feature embedding}
    \label{fig:mod4}
\end{figure}

Fourier feature embedding maps low-dimensional inputs, such as spatial coordinate $ z $ and time $ t $ in our case, into a higher-dimensional space using sinusoidal functions of multiple frequencies. Formally, each input is transformed via:

\begin{equation}\label{eq:ffe}
    \gamma (\textbf{c}) = \begin{bmatrix}
    \sin{(2\pi \mathbf{B}\textbf{c}}) \\
    \cos{(2\pi \mathbf{B}\textbf{c}})
\end{bmatrix}^T 
\end{equation}

where $\textbf{c}$\nomenclature{$\textbf{c}$}{The input feature vector for the Fourier feature embedding function} is the input feature vector (trunk net input in this problem) of length $k_f$\nomenclature{$k_f$}{The length of the input feature vector $\textbf{c}$ for Fourier embedding}, and  $\gamma (\textbf{c})$ is the embedded Fourier feature vector with a length of $2m_f$. $\mathbf{B}$\nomenclature{$\mathbf{B}$}{The matrix of frequencies used in Fourier feature embedding} is a $m_{f}$\nomenclature{$m_{f}$}{The hyperparameter for Fourier embedding size} by $k_{f}$ matrix of frequencies, sampled from $\mathcal{N}(\mathbf{0},\sigma_f^2I)$. The tunable hyperparameters $m_f$ and $\sigma_f$\nomenclature{$\sigma_f$}{The standard deviation for sampling the frequency matrix $B$} are pre-specified before the neural network training. This embedding introduces periodic features that enrich the input representation, allowing neural networks to approximate functions with high variation. Leveraging the Fourier feature embedding, we propose the following architecture:

\begin{itemize}
    \item \textbf{Model 4 (\Cref{fig:mod4}):} The DeepONet architecture is the same as Model 3; however, the Fourier feature embedding maps the input $(z, t, C_v)$ into a higher-dimensional latent space before it is passed to the trunk net. This embedding transforms the original low-dimensional input into a richer representation using sinusoidal functions of varying frequencies, enabling the trunk net to learn functions with large variations. 
\end{itemize}

All the networks (i.e., branch, trunk, and merge net) in DeepONet models in this study employ fully connected multilayer perceptrons (MLPs). Each MLP comprises 6 layers with 30 neurons per layer. The output dimension of the branch and trunk net for the dot product is set to $q=50$ ( \Cref{eq:deeponet_architecture}). For Fourier feature embedding hyperparameters (\Cref{eq:ffe}) in Model 4, \added{we performed a randomized search to select $m_f=50$ and $\sigma_f=1.0$. Previous studies (e.g., \cite{tancik2020fourier,liu2025leveraging}) have conducted ablation studies to evaluate the impact of the Fourier feature parameters on the performance of neural networks. Specifically, increasing the value of $m_f$ allows neural networks to capture a broader range of frequency components, though this comes at the cost of more computational time.  The parameter $\sigma_f$ controls the frequency range covered by the Fourier features in the neural tangent kernel. A larger $\sigma_f$ allows for the inclusion of higher frequency components, which helps the neural network learn finer details. Conversely, a smaller $\sigma_f$ restricts the frequency range, resulting in overly smooth representations that may miss important details.} All the DeepONet models are trained using the AdamW optimizer with an initial learning rate of 0.01 and a mini-batch size of 256.

\section{Results}\label{sec:results}
This section presents the performance evaluation of the four investigated DeepONet architectures in predicting excess PWP dissipation for the consolidation problem. We assess the models based on their accuracy under varying initial excess PWP $u_0$ and coefficient of consolidation $C_v$ conditions, and their computational efficiency.

% Typical results with non-uniform u0:

\begin{figure}[!htbp]
    \centering
    \includegraphics[width=1.0\textwidth]{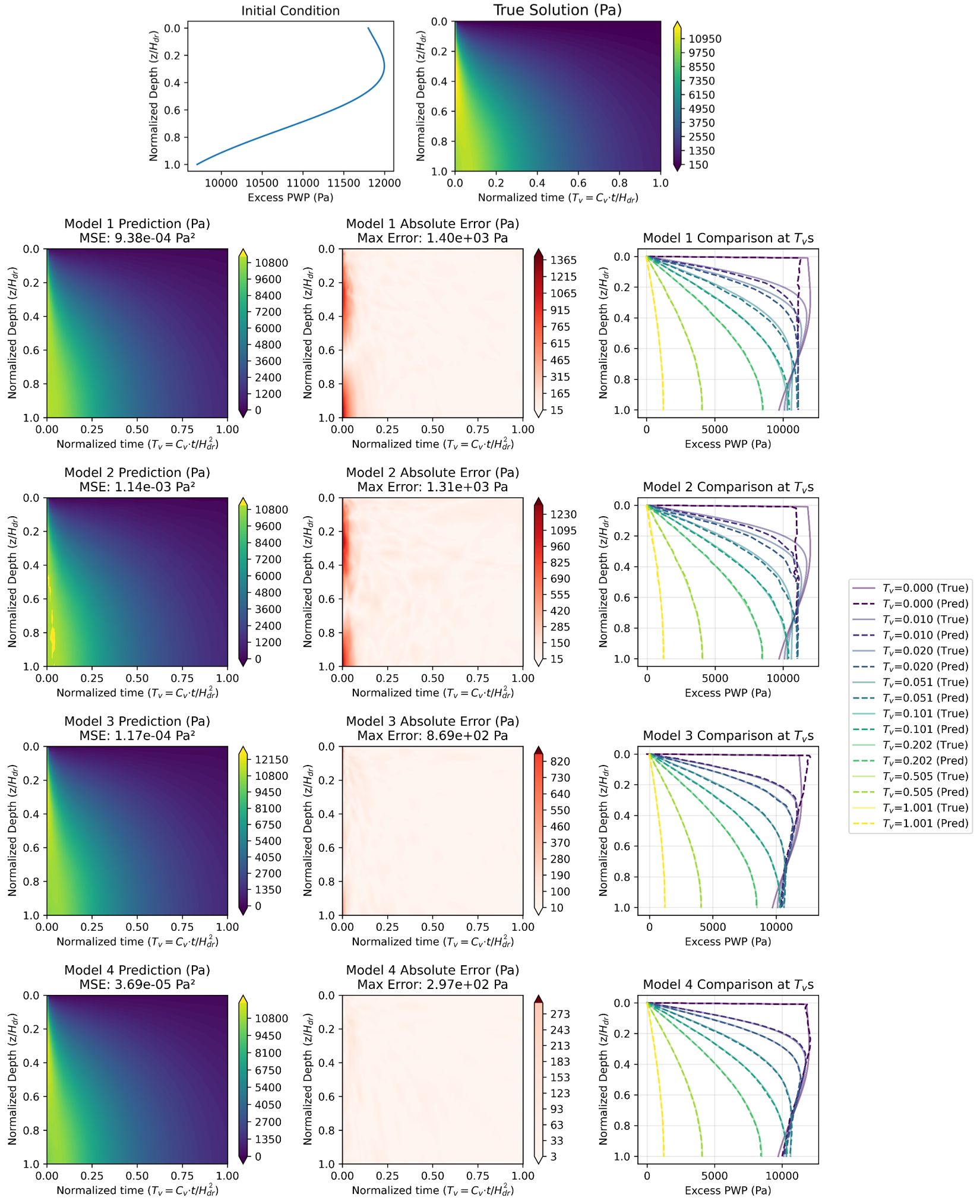}
    \caption{Excess pore water pressure (PWP) dissipation evaluated by the different DeepONet architectures on test data at $C_v=0.5 \ m^2/year$ with varying initial excess PWP over the depth. The figures in the first row show the initial excess PWP function and the true solution. The subsequent rows show the prediction from each model with the absolute error compared to the true solution, and the plots comparing the true and predicted excess PWP distributions over depth at selected times.}
    \label{fig:results_varing_u0}
\end{figure}

\Cref{fig:results_varing_u0} shows the comparative performance of the four DeepONet architectures for a representative case in the test data with a non-uniform $u_0$ and a $C_v=0.5 \ m^2/year$. A non-uniform $u_0$ is selected as it imposes fluctuating responses early in the consolidation process and enables a more rigorous assessment of the implemented architectures. Moreover, non-uniform $u_0$ profiles appear in geotechnical practice \citep{lovisa2015non_uniform_u0}, for instance, due to staged construction, uneven surface loading, or localized drainage. 

The first row of \Cref{fig:results_varing_u0} shows the initial excess PWP function and the true solution. After this row, each of the figures corresponds to models 1 to 4. For each model, three subplots are presented: (1) the predicted spatiotemporal excess PWP field with MSE value in the figure title, (2) the corresponding absolute error with maximum error value in the title, and (3) a comparison of predicted and true pore pressure profiles at selected time snapshots.

All models capture the overall spatiotemporal excess PWP dissipation trends; however, the accuracy varies depending on the models and the region. Model 1, the standard DeepONet with $C_v$ concatenated to the initial excess PWP in the branch net, yielded a MSE of $9.38\times10^{-4} \ Pa^2$ and a maximum absolute error of $1.40\times10^{3} \ Pa$. Model 2, which processes $C_v$ in a separate branch net before merging, exhibits a higher MSE of $1.14\times10^{-3} \ Pa^2$ and a maximum absolute error of $1.31\times10^{3} \ Pa$. While both models capture the overall spatiotemporal trend, they display noticeable deviations from the true solution and larger errors during the early stages of consolidation, particularly in regions with the steepest pore pressure gradients.

Model 3, where we incorporate $C_v$ as an input to the trunk net, shows a significant improvement in prediction accuracy. This architecture achieves an MSE of $1.17\times10^{-4} \ Pa^2$ and a maximum absolute error of $8.69\times10^{2} \ Pa$. The predictions from Model 3 align more closely with the true solution across the spatiotemporal domain, with substantially reduced absolute errors, especially in areas of rapid excess PWP change. However, slight elevations in error are still present at the very initial part of the consolidation process for this model.

Model 4, which augments Model 3 by applying Fourier feature embedding to the trunk net's input $(z, t, C_v)$, demonstrates the highest accuracy. It achieves the lowest MSE of $3.69\times10^{-5} \ Pa^2$ and the smallest maximum absolute error of $2.97\times10^{2} \ Pa$ for this case. Model 4 effectively minimizes the errors that the other models show, particularly reducing errors at the early stages of consolidation where excess PWP pressure changes are most pronounced.

In the time snapshot plots (rightmost subplots), early-stage prediction errors become more evident. Models 1 and 2 exhibit large discrepancies in the pore pressure distribution during the initial phase (from $ T_v = 0.0 $ to $ 0.028$), not accurately capturing the rapid variation in pore pressure. In contrast, Models 3 and 4 more accurately reproduce the early-time profiles. However, Model 3 still shows a slight mismatch at $ T_v = 0.0 $. This residual error is effectively mitigated by Model 4, which demonstrates improved accuracy at the earliest time step, owing to the enhanced trunk net input representational capacity introduced by the Fourier feature embedding.

\begin{figure}[!htbp]
    \centering
    \includegraphics[width=1.0\textwidth]{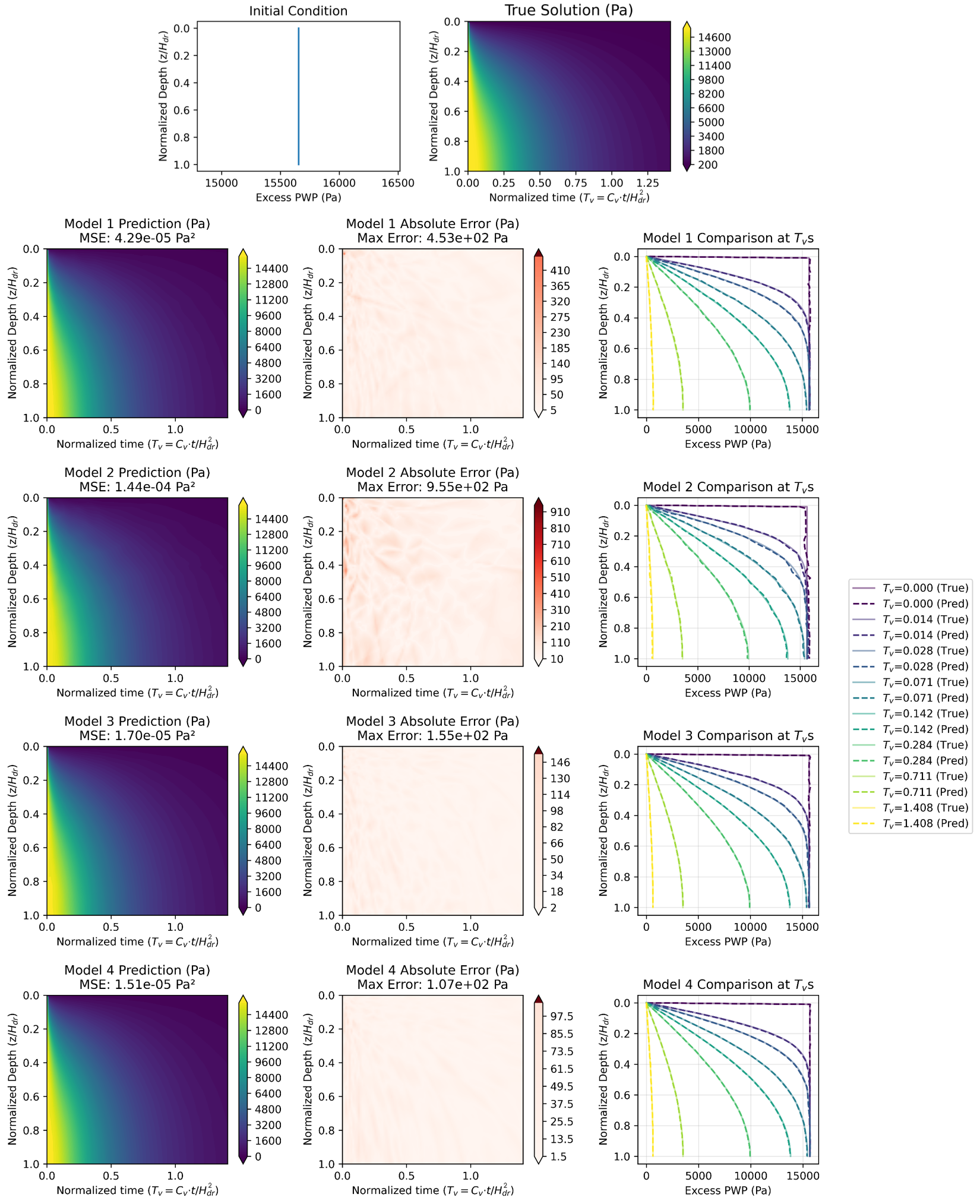}
    \caption{Excess pore water pressure (PWP) dissipation evaluated by the different DeepONet architectures on test data at $C_v=0.7\ m^2/year$ with constant initial excess PWP over the depth. The figures in the first row show the initial excess PWP and the true solution. The subsequent rows show the prediction from each model with the absolute error compared to the true solution, and the plots comparing the true and predicted excess PWP distributions over depth at selected times.}
    \label{fig:results_constant_u0}
\end{figure}

% Typical results with uniform u0:
\Cref{fig:results_constant_u0} presents the model performances for a case in the test data with a uniformly distributed initial excess PWP with $C_v=0.7 \ m^2/year$. The configuration of the figure is the same as \Cref{fig:results_varing_u0}. In this simpler case, overall errors tend to be smaller than those from the non-uniform case shown in \Cref{fig:results_varing_u0}. The concentration of errors near the early stage of consolidation is similarly evident as in the case shown in \Cref{fig:results_varing_u0}. 

Model 1 achieves an MSE of $4.29\times10^{-5} \ Pa^2$, with moderate absolute errors predominantly concentrated near the early time steps. Model 2 again shows the weakest performance, recording the highest MSE of $1.44\times10^{-4} \ Pa^2$ and the largest maximum absolute error of 955 Pa. 

Model 3, with $C_v$ in the trunk net, again demonstrates a clear advantage, yielding an MSE of $1.70\times10^{-5} \ Pa^2$ and a maximum absolute error of $1.55\times10^{2} \ Pa$. Its predictions accurately track the time-dependent behavior with reduced absolute error. Model 4 achieves the highest accuracy by further improving upon Model 3 through the incorporation of Fourier feature embedding. Model 4 results in the lowest MSE of $1.51\times10^{-5} \ Pa^2$ and the smallest maximum error of $107 Pa$. 

In the time snapshot plots (rightmost subplots), pressure distributions are mostly accurate for all models, including the early stages. However, we observe a few fluctuations at $T_v$ = 0.0 for Model 1 and 2, and this fluctuation extends to $T_v = 0.028$. Model 4 accurately captures all the early pressure distributions.

\begin{figure}[!htbp]
    \centering
    \includegraphics[width=1.0\textwidth]{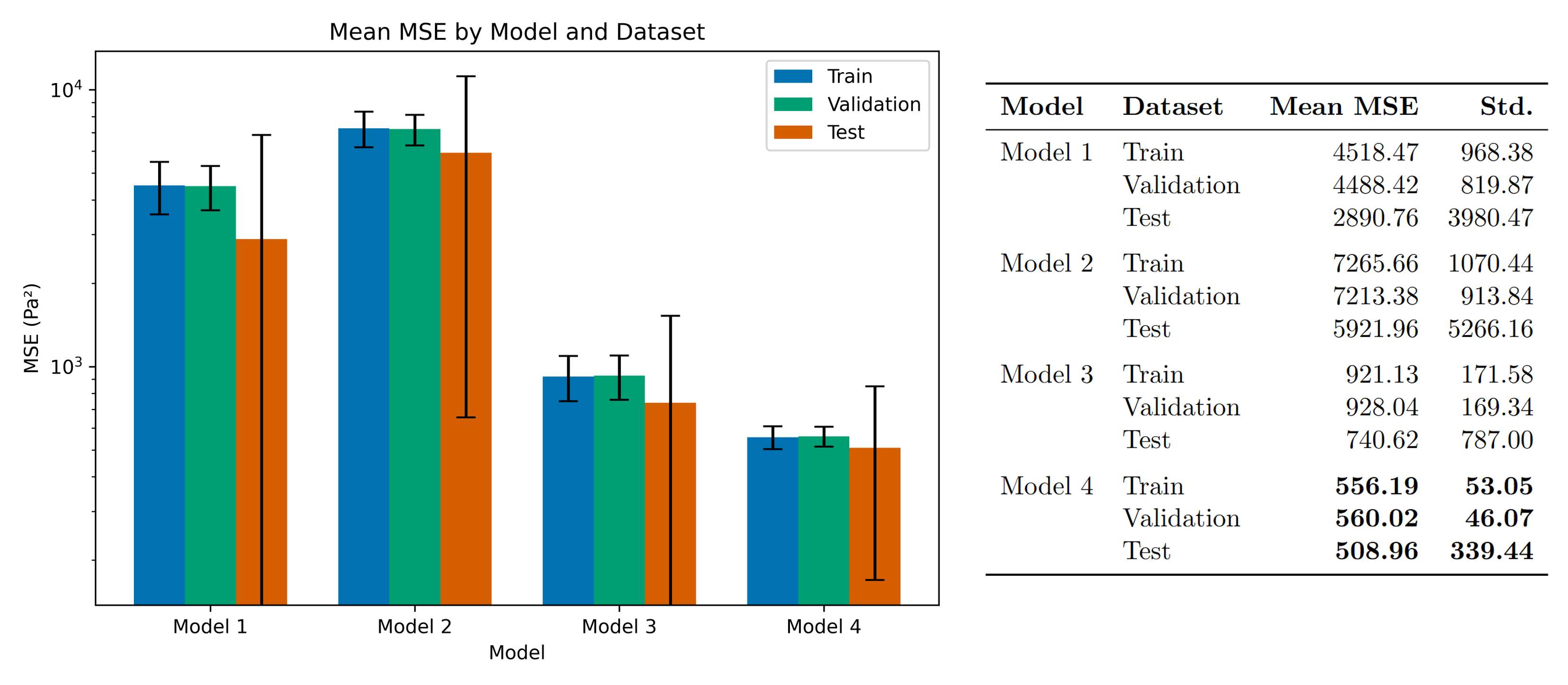}
    \caption{Mean and standard deviation of MSE for each DeepONet architecture across training, validation, and test datasets in physical scale.}
    \label{fig:mean_mse}
\end{figure}

% Model statistics
\Cref{fig:mean_mse} shows the mean and standard deviation of MSE for each DeepONet architecture across training, validation, and test datasets, with a table showing the corresponding values. 
Models 1 and 2 consistently exhibit the highest MSEs and larger standard deviations. Model 2, featuring separate branch networks for the initial condition and $C_v$ performs the poorest, with a test MSE of 5921.96±5266.16 $Pa^2$ (mean ± 1 standard deviation). Model 3 significantly reduces the MSE across all datasets, achieving a test MSE of 740.62 ± 787.00 $Pa^2$, which is about 5 to 7 times lower than models 1 and 2. Model 4 shows the most accurate and consistent performance. The application of Fourier feature embedding to the trunk input $(z, t, C_v)$ in Model 4 results in the lowest test MSE values, recording a test MSE of 508.96 ± 339.44 $Pa^2$, which further reduces the MSE of the model by about 1.5 times, along with markedly reduced variance compared to other models.

\begin{figure}[!htbp]
    \centering

    \begin{subfigure}[b]{1.0\textwidth}
        \centering
        \includegraphics[width=\textwidth]{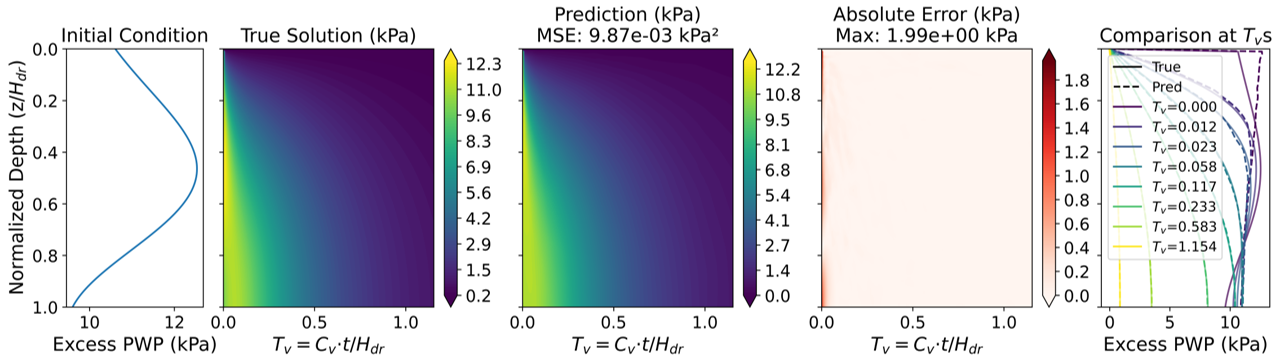}
        \caption{Model 3 ($C_v=0.5771 \ m^2/year$)}
        \label{fig:model3worst}
    \end{subfigure}
    \vfill
    \begin{subfigure}[b]{1.0\textwidth}
        \centering
        \includegraphics[width=\textwidth]{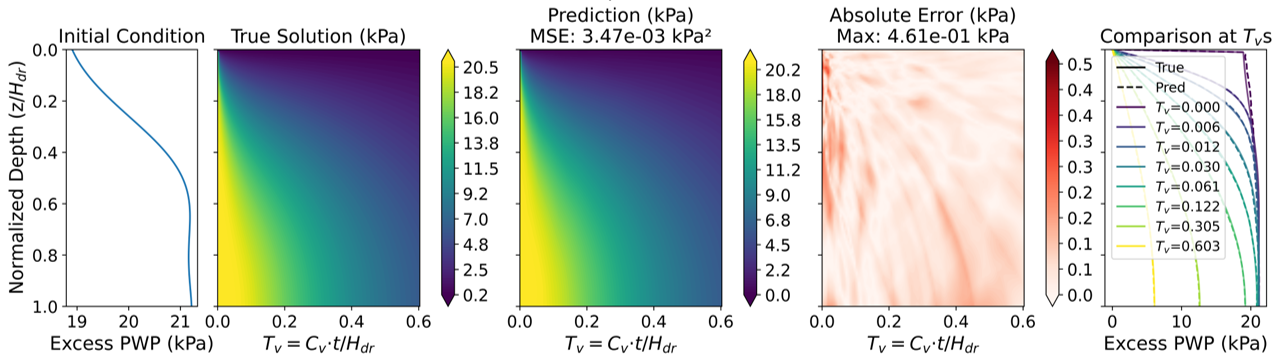}
        \caption{Model 4 ($C_v=0.3016 \ m^2/year$)}
        \label{fig:model4worst}
    \end{subfigure}

    \caption{Worst case predictions for Model 3 and Model 4. The plots share the y-axis. The figures do not share the same color bar scale for the prediction and error fields.}
    \label{fig:modelsworst}
\end{figure}

To further investigate the impact of Fourier feature embedding, we compare the worst-case predictions of the best-performing architectures: Model 3 and Model 4. These cases, shown in \Cref{fig:modelsworst}, correspond to the samples with the highest mean squared error (MSE) for each respective model. Note that the color bars for the prediction and absolute error fields differ between the two models to reflect their distinct value ranges. 

Model 4 demonstrates superior performance even in its worst-case scenario, with an MSE of 3.47e+03 $Pa^2$, approximately three times lower than the worst-case MSE of 9.87e+03 $Pa^2$ of Model 3. A closer examination of the spatiotemporal absolute error distributions reveals key differences. It shows that Model 3 exhibits highly localized error spikes, particularly at early times characterized by rapid excess PWP dissipation and high variability. The temporal comparison plots (rightmost column in each figure) for Model 3 show difficulty in matching the true pressure profiles during the initial phase ($T_v$=0.0 to 0.012). In contrast, Model 4 maintains a more uniformly distributed error profile with a generally reduced magnitude across the domain and achieves closer alignment with the true solution at all selected time snapshots, even in this worst-case prediction.

\begin{figure}[!htbp]
    \centering
    \includegraphics[width=1.0\textwidth]{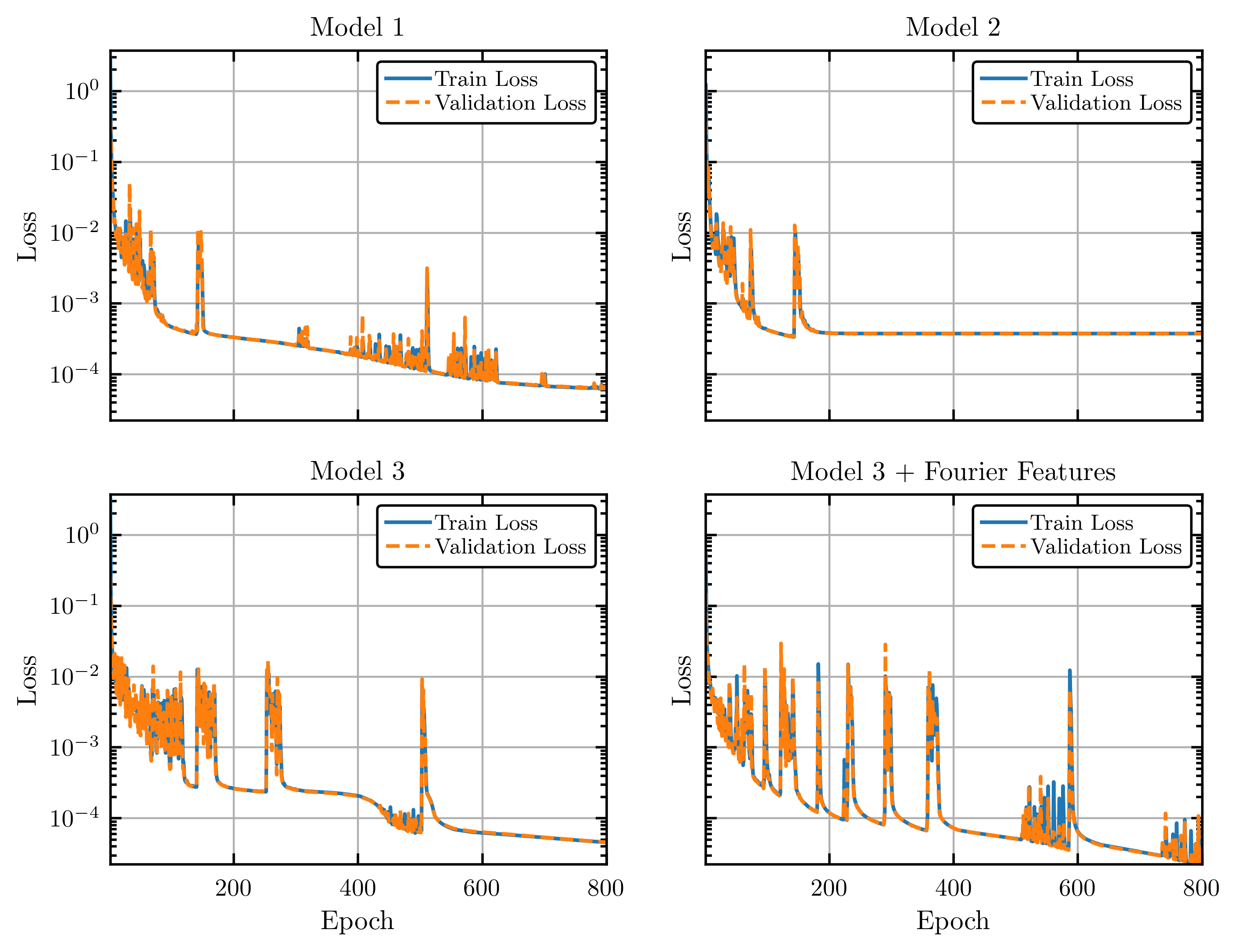}
    \caption{Training and validation loss curves for DeepONet architectures.}
    \label{fig:loss_curve}
\end{figure}

\Cref{fig:loss_curve} presents the training and validation loss curves for the four DeepONet architectures. All models show a consistent alignment between training and validation curves, indicating the generalization and effectiveness of DeepONet's solution operator learning. However, each model achieves different loss levels. Model 1 and 2 demonstrate gradual loss reduction with mild fluctuations, but their convergence stabilizes at a relatively higher loss level. Model 2 converges at around 200 epochs, but this stabilization does not translate into strong predictive performance. Model 3 achieves notably lower losses and tends to keep decreasing even after 600 epochs, reflecting its superior performance. Model 4, while displaying some oscillations during early epochs, likely due to the increased complexity from Fourier feature embedding, ultimately attains the lowest overall losses, confirming its enhanced expressive capacity and improved accuracy.

Oscillatory patterns in the loss histories \Cref{fig:loss_curve} are often observed in DeepONet and other deep learning models (e.g., supplementary materials from \cite{lu2021learning} and \cite{wang2021learning_pideeponet}). These oscillations can be attributed to the stochastic nature of mini-batch gradient updates in the AdamW optimizer, particularly when using relatively small batch sizes and learning rates. Such fluctuations reflect stochastic variations preventing convergence to non-optimal local minima \citep{lu2023benign_lr_large}. In general, adjusting the batch size or learning rate, or applying a learning-rate decay schedule, could help smooth the training curves \citep{smith2017don_decay_lr}. In this study, training was stopped at approximately 800 epochs due to GPU runtime constraints on the computing cluster, and the model at this epoch was used for consistent comparisons across architectures. \added{While no explicit early-stopping criterion was used, we monitored training and validation losses throughout training to assess convergence. Convergence was evaluated from the validation-loss trajectory. Across all models, improvements diminish in later epochs, and we observe no systematic separation between training and validation losses or evidence of significant loss instability.}

The training time for models 1 to 4 is approximately 32, 44, 31, 35 minutes using NVIDIA RTX 4090 GPU. The elevated training time for Model 2 reflects the additional network introduced for the merge net. The additional Fourier feature embedding operation in Model 4 did not increase the training time significantly.

% Talk about computation speed
\Cref{tab:comp_time} provides a comparison of the average computation times to generate a full spatiotemporal excess PWP solution field for the four DeepONet architectures and two classical numerical solvers (implicit BDF and explicit Runge-Kutta RK45). The reported times are averaged over 500 test cases, each corresponding to a distinct initial excess PWP profile (i.e., input function), and reflect the time to compute the entire solution field per case on an NVIDIA RTX 2000 Ada Laptop GPU. All DeepONet models demonstrate rapid prediction capabilities, with average execution times ranging from approximately 0.006 to 0.008 seconds per case. Model 4 is the fastest DeepONet architecture with a mean prediction time of 0.00619 s, followed by Model 3 (0.00677 s), Model 1 (0.00709 s), and Model 2 (0.00822 s).

Comparatively, the implicit BDF solver requires an average of 0.01118 s to produce a solution field under the same conditions used to evaluate DeepONet computation time, while the explicit RK45 solver is significantly slower, averaging 0.71304 s. Thus, the trained DeepONet models are approximately 1.5 to 2 times faster than the implicit numerical solver and about 100 times faster than the explicit numerical solver. 

While the absolute time differences are marginal (order of milliseconds to sub-seconds) for 1D consolidation (\cref{tab:comp_time}), the results emphasize DeepONet’s potential as a computationally efficient surrogate model. This efficiency could be particularly significant when modeling large-scale or high-dimensional problems where conventional solvers may be computationally costly. Once trained, DeepONet enables rapid inference at any queried spatiotemporal regions without time-stepping across varying initial and material conditions and without retraining. 

\begin{table}[ht]
\centering
\caption{Comparison prediction/execution time (in seconds) for DeepONet architectures and classical numerical solvers for 1D case.}
\begin{tabular}{@{}lcc@{}}
\toprule
\textbf{Method} & \textbf{Mean (s)} & \textbf{Std (s)} \\
\midrule
Model 1 (Standard) & 0.00709 & 0.00281 \\
Model 2 ($C_v$ Branch) & 0.00822 & 0.00316 \\
Model 3 ($C_v$ Trunk) & 0.00677 & 0.00187 \\
Model 4 ($C_v$ Trunk + Fourier) & \textbf{0.00619} & \textbf{0.00166} \\
Implicit solver (BDF) & 0.01118 & 0.00095 \\
Explicit solver (RK45) & 0.71304 & 0.11409 \\
\bottomrule
\end{tabular}
\label{tab:comp_time}
\end{table}

\subsection{Application to 3D Consolidation}

%%% \added: Briefly mention 3D PDE------------------------------------------------
We further extend the application of Model 4 to 3D consolidation scenarios; the motivation is to have a scalability assessment. The governing equation for 3D consolidation is:

\begin{equation}
    \frac{\partial u}{\partial t}
    = C_v \left(
    \frac{\partial^2 u}{\partial x^2}
    + \frac{\partial^2 u}{\partial y^2}
    + \frac{\partial^2 u}{\partial z^2}
    \right)
\label{eq:3D_consol_pde}
\end{equation}

Here, $u$ is defined in a 3D space with $u(x,y,z,t)$ at time $t$. In this case, fully drained boundary conditions ($u=0$) are applied on all faces of a cubic domain. %A constant and isotropic coefficient of consolidation, $C_v$, is used}

The numerical formulation for the 3D case follows the same finite difference approach used in the 1D consolidation problem. The domain is uniformly discretized into $N_x$, $N_y$, and $N_z$ nodes along the $x$, $y$, and $z$ directions, respectively, with grid spacings $\Delta x$, $\Delta y$, and $\Delta z$. Applying a second-order central finite-difference approximation in each direction, \Cref{eq:3D_consol_pde} can be expressed for an internal node $(i,j,k)$ as:

\begin{equation}
    \frac{du_{i,j,k}}{dt} = C_v \left(
    \frac{u_{i+1,j,k} - 2u_{i,j,k} + u_{i-1,j,k}}{\Delta x^2} +
    \frac{u_{i,j+1,k} - 2u_{i,j,k} + u_{i,j-1,k}}{\Delta y^2} +
    \frac{u_{i,j,k+1} - 2u_{i,j,k} + u_{i,j,k-1}}{\Delta z^2}
    \right)
    \label{eq:3D_fdm}
\end{equation}

Similar to the 1D case, we use RK methods to solve the discretized ODE system using the \texttt{solve\_ivp} function from the SciPy library %to generate data.

%%% \added: training data------------------------------------------------

% N functions, u0, P, Cv, dataset

\added{We generate N = 20,000 and N = 2,000 functions and coefficients for training and validation, respectively, and test the trained model on 100 unseen functions and coefficients, with no overlap between the three sets.} Specifically, the initial excess pore water pressure, $u(x,y,z,0)$, varies spatially in the $x$ and $y$ directions and is modeled using GRF, as in the 1D case. The GRF is defined with $\sigma_r^2 = 1 \ \text{kPa}^2$ and $l = 0.30$ in $x$ and $y$ directions. $\mu_r$ is uniformly sampled from 10 to 20 kPa. The input function is represented by $m = 50 \times 50$ equally spaced locations distributed over the $x$-$y$ plane (i.e., $\{x_{\ell}\}_{\ell}^{m=50\times50}$). The corresponding solutions are sampled at P=10,000 locations uniformly, (i.e., $ \{y_j^{(i)}\}_{j=1}^{10,000} $), in the four-dimensional space $(x, y, z, t)$. As a result, we create $20,000 \times 10,000$ training data. $C_v$ is sampled uniformly from the range $0.3$-$1.0 \ m^2/year$. The validation set is constructed in the same manner, with P=10,000 output evaluations per input function, resulting in $ 2,000 \times 10,000 $ pairs. The spatial domain spans the normalized $x$, $y$, $z$ and $t$, i.e., $\frac{x, \ y, \ z}{H_{dr}} \in [0, 2]$  and normalized time $T_v \in [0, 1]$.

% \added: add 3D deeponet structure ------------------------------------------------
The branch and trunk net uses MLPs with 5 layers and 128 neurons per layer. Their output dimensions for the dot product are set to 128. The Fourier feature embedding uses $m_f=50$ and $\sigma_f=1.0$. We use the AdamW optimizer, with a learning rate of 1.0e-4, a batch size of 16,834. Compared to the 1D case in Model 4, this setting uses a larger MLP and a larger batch size during training to achieve higher data throughput. We train the model for 70 epochs, taking about 11 hours using the same GPU used in the 1D case.

\begin{figure}[!htbp]
    \centering
    \includegraphics[width=1.0\textwidth]{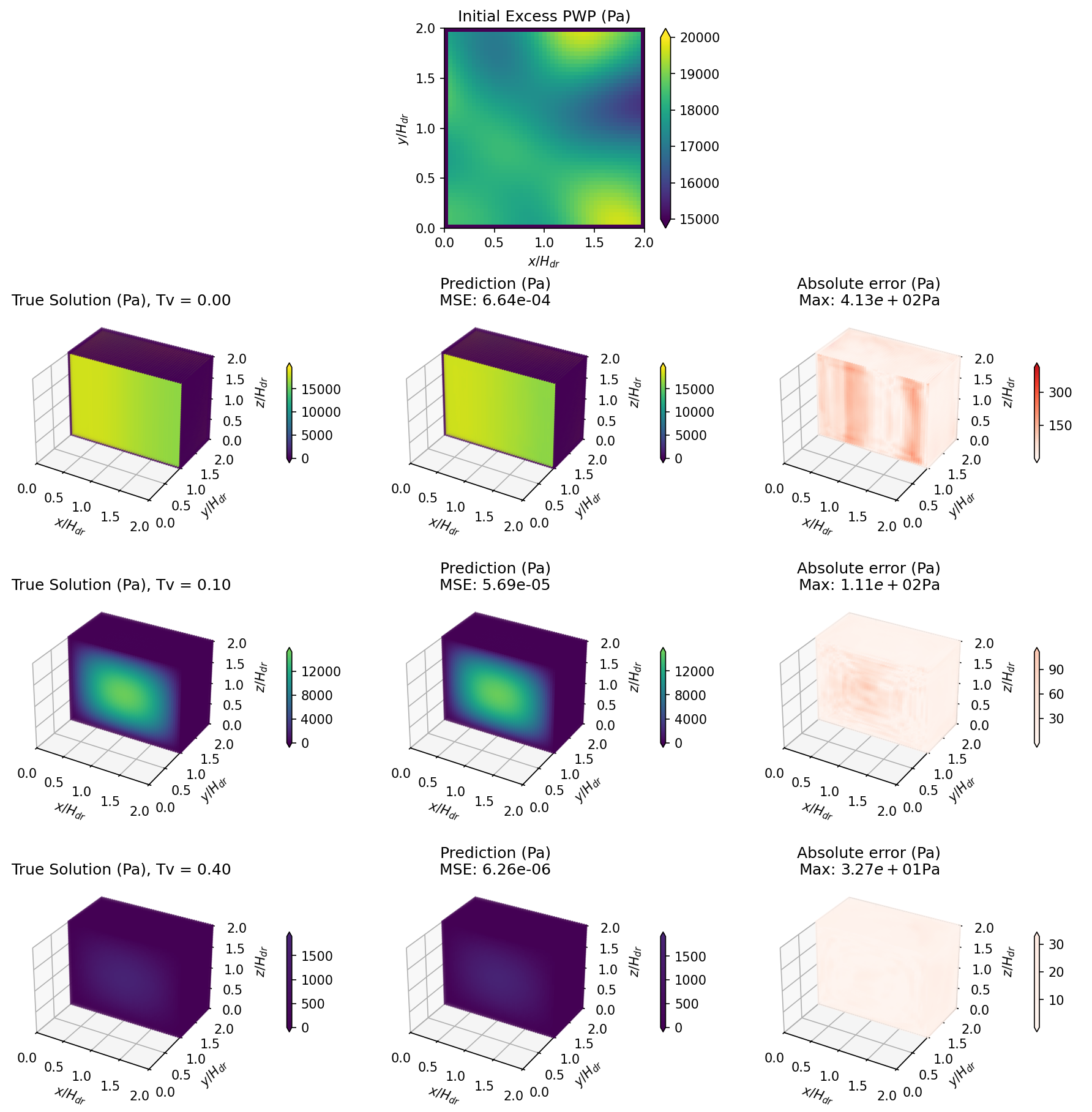}
    \caption{Excess PWP dissipation evaluated by the 3D DeepONet architectures on test data at $C_v=0.5 \ m^2/year$ with spatially varying initial excess PWP under full drainage condition. The subsequent rows show the solution field from the finite difference method, prediction from DeepONet, and the absolute error, at the specified $T_v$}
    \label{fig:three_d_result}
\end{figure}

%%% \added: results------------------------------------------------
We evaluate the learned solution operator $\mathcal{G}_{\Theta}(u^{(i)})(y)$ at a $50^4$ uniformly spaced grid of $y(x,y,z,t)$ points across the output domain. \Cref{fig:three_d_result} shows the result with $C_v=0.50 \ m^2/year$ at $T_v=0, \  0.10, \ 0.40$ (results for other $C_v$ are similar). The first row shows the input function $u_0$. DeepONet captures the overall spatio-temporal PWP dissipation response successfully with an average MSE of 1.07e+03 $Pa^2$. The prediction shows the largest overall error at the very first timestep with the maximum absolute error of 4.13e+2 $Pa$. Compared to the 1D case in \Cref{fig:results_varing_u0}, this value is slightly larger than Model 4's maximum absolute error (2.97e+2 $Pa$) and smaller than the maximum absolute error from Model 3 (8.69e+2 $Pa$). For the later times, the prediction errors become almost negligible.

\begin{figure}[!htbp]
    \centering
    \includegraphics[width=1.0\textwidth]{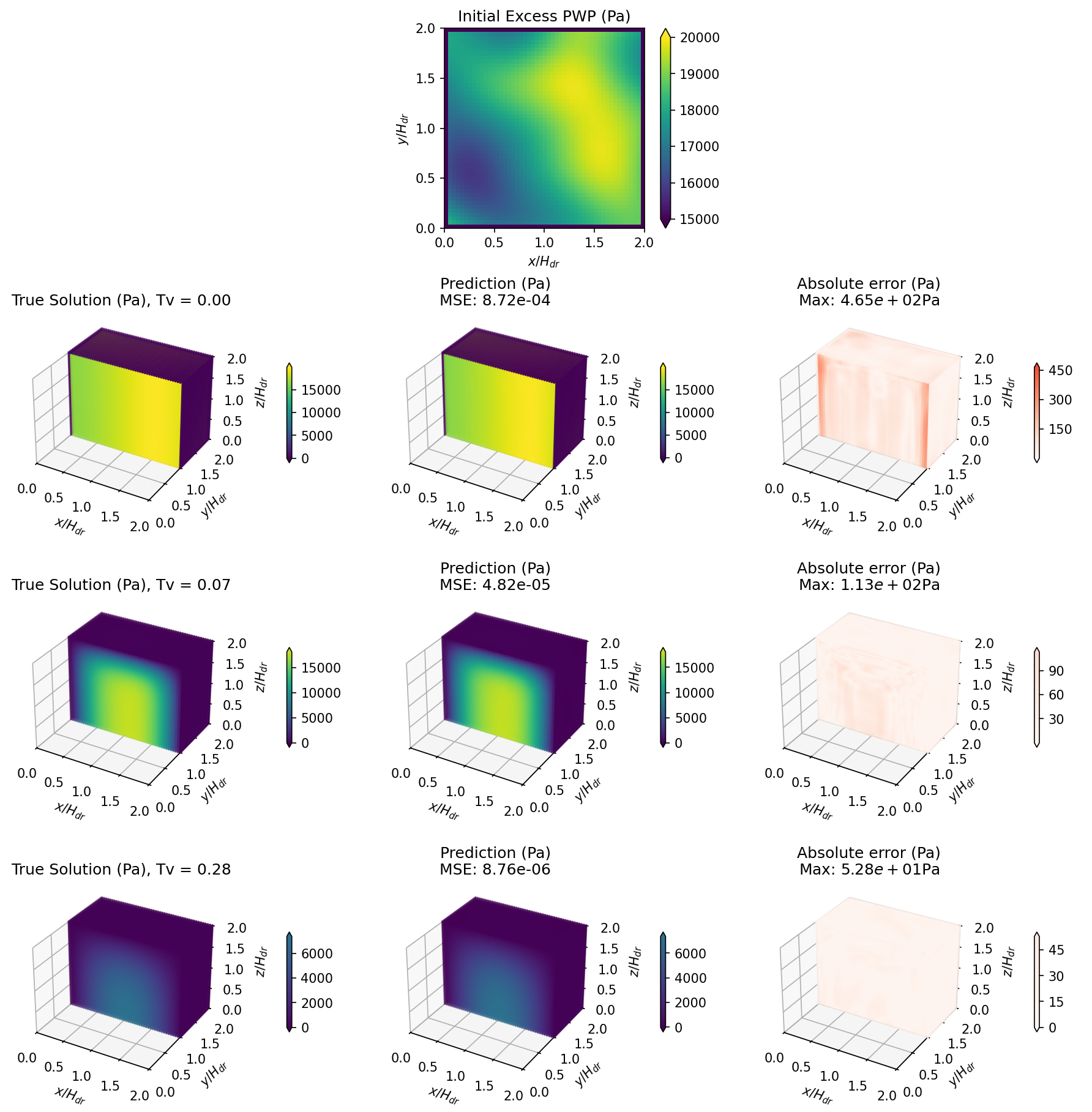}
    \caption{Excess PWP dissipation evaluated by the 3D DeepONet architectures on test data at $C_v=0.35 \ m^2/year$ with spatially varying initial excess PWP under bottom impermeable condition. The subsequent rows show the solution field from the finite difference method, prediction from DeepONet, and the absolute error, at the specified $T_v$}
    \label{fig:three_d_result_different_bc}
\end{figure}

\added{
DeepONets are not restricted to the specific boundary conditions and can be expanded to mixed boundary conditions depending on training data coverage. We prepare an additional consolidation dataset with a bottom impermeable condition with the same data size as the full drainage case to train the model. \Cref{fig:three_d_result_different_bc} shows an evaluation result with $C_v=0.35 \ m^2/year$ at $T_v=0, \  0.07, \ 0.28$. Similar to the case with full drainage conditions, the prediction shows a good agreement with the finite difference solver. Although we do not address the other boundary conditions, in principle, they can be accommodated by incorporating boundary-condition indexing variables or additional input channels into the operator formulation.
}

% Talk about train/val/test MSE compared to 1D case.
\added{\Cref{table:mean_mse_3d} shows the mean and standard deviation of MSE for evaluating the full solution field in physical scale across training, validation, and test datasets for both full drainage and bottom impermeable boundary conditions. We can see that these MSEs span between the values of Model 3 and Model 4 in the 1D case (see \Cref{fig:mean_mse}), with significantly smaller values than Model 1 and 2. Given the greater complexity and larger sampling space of the 3D problem compared to the 1D case, Model 4 demonstrates good performance across the considered 3D scenarios.}

\begin{table}[htbp]
\centering
\caption{Model performance under different drainage conditions.}
\label{table:mean_mse_3d}
\begin{tabular}{lcccc}
\toprule
\multirow{2}{*}{Dataset} 
& \multicolumn{2}{c}{Full Drainage} 
& \multicolumn{2}{c}{Bottom Impermeable} \\
\cmidrule(lr){2-3} \cmidrule(lr){4-5}
& Mean MSE ($\mathrm{Pa}^2$) & Std (Pa) 
& Mean MSE ($\mathrm{Pa}^2$) & Std (Pa) \\
\midrule
Training   & $7.65\times10^{2}$ & $3.01\times10^{1}$ & $5.49\times10^{2}$ & $2.45\times10^{1}$ \\
Validation & $1.09\times10^{3}$ & $2.90\times10^{1}$ & $5.49\times10^{2}$ & $2.62\times10^{1}$ \\
Test       & $7.78\times10^{2}$ & $3.51\times10^{2}$ & $5.13\times10^{2}$ & $2.61\times10^{2}$ \\
\bottomrule
\end{tabular}
\end{table}

\Cref{table:comp_time_3d} compares the computation time required to generate a full 3D solution field between the trained DeepONet and the explicit numerical solver (RK45). DeepONet predicts the complete 3D field almost instantaneously (about 0.1 s), whereas the classical solver requires over 120 s on average, achieving a speedup of more than three orders of magnitude. The implicit BDF solver was unable to produce results, as the computer memory ran out due to the prohibitively large Jacobian matrix in the 3D configuration. 

%% \added: add 3D computation time comparison table
\begin{table}[ht]
\centering
\caption{Comparison of prediction/execution time (in seconds) for DeepONet and classical numerical solvers for the 3D case.}
\begin{tabular}{@{}lcc@{}}
\toprule
\textbf{Method} & \textbf{Mean (s)} & \textbf{Std (s)} \\
\midrule
Model 4 (Standard) & 0.101 & 0.019 \\
Explicit solver (RK45) & 126.63& 4.30\\
\bottomrule
\end{tabular}
\label{table:comp_time_3d}
\end{table}

\begin{figure}[!htbp]
    \centering
    \includegraphics[width=1.0\textwidth]{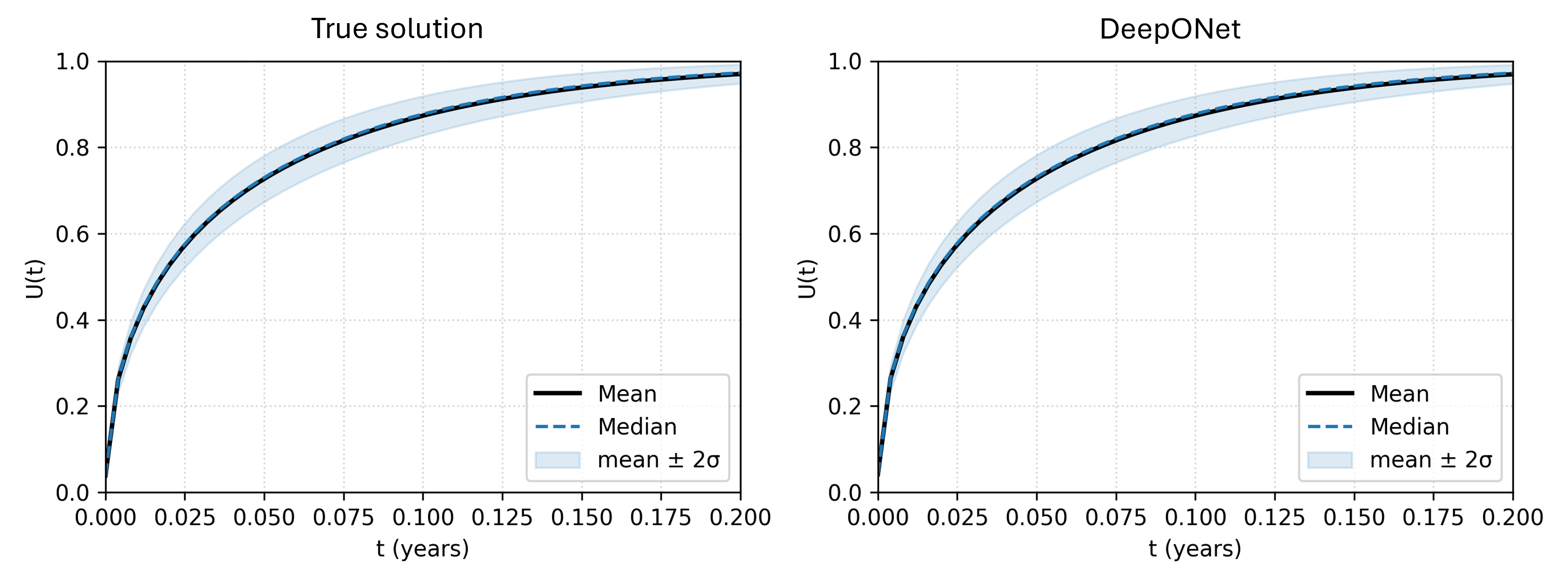}
    \caption{Stochastic evolution of the degree of consolidation $U(t)$ obtained from the finite difference solver (left) and DeepONet (right), showing close agreement in mean trends and $\pm 2\sigma$ uncertainty bounds.}
    \label{fig:uq_3d}
\end{figure}

\Cref{fig:uq_3d} further illustrates the performance of the trained DeepONet in 3D scenarios by showing the evolution of the spatially averaged degree of consolidation $U(t)$ under uncertain conditions, comparing DeepONet predictions with those from the finite difference solver (reference solution). \Cref{fig:uq_3d} also highlights a potential practical application of DeepONet because $U(t)$ is a key engineering parameter to make decisions, and quantifying its uncertainty is of interest. In engineering applications, pore pressure instrumentation often reveals spatially variable piezometric regimes \citep{tang2025consol_non_uniform_loads, darrag1993consolidation_stochastic_initial}, which can be approximated by random fields using geostatistical techniques (e.g., \cite{macedo2024cptu}), making them well suited for DeepONet-based surrogate modeling. Moreover, engineering assessments commonly use mean values of $C_v$, varied over lower and upper bounds to reflect parameter uncertainty \citep{macedo2024cptu}.  We generate 1,000 realizations of the initial excess PWP using GRF with $\sigma_r=1 \ kPa^2$, $\mu_r=15 \ kPa$, and $l=0.3$. $C_v$ is sampled from a normal distribution with mean $0.5~\mathrm{m^2/year}$ and a coefficient of variation of 0.1. For each realization, the corresponding $U(t)$ is computed by using the dissipated pore pressures at a given time and averaging over the calculation domain; this process is described in classical geotechnical books (e.g., \cite{holtz1981introduction_geotech}). The $U(t)$ variations at a given time are used to approximate its statistical distribution and plot the uncertainty range in \Cref{fig:uq_3d}. DeepONet accurately reproduces uncertainty propagation, yielding $\pm 2\sigma$ bounds nearly identical to those from the finite difference solver while retaining the computational efficiency reported in \Cref{table:comp_time_3d}. For example, at $t = 0.1$~years, the mean and standard deviation of $U$ from the numerical solver are 0.8733 and 0.0227, respectively, whereas DeepONet yields 0.8731 and 0.0228, demonstrating an excellent quantitative agreement. These results indicate that DeepONet can efficiently and reliably quantify uncertainty under variable input conditions, offering a promising surrogate modeling strategy for complex consolidation problems. In practice, DeepONet could be integrated with modern sensing technologies (e.g., \cite{anjana2025fiber_pwp} and \cite{hottges2023fiber_novel}) that collect dense pore-pressure measurements, together with robust subsurface characterization methods to estimate bounds on $C_v$. Such a framework has the potential to enable computationally efficient, large-scale, near real-time consolidation monitoring, where DeepONet’s speed offers substantial advantages over traditional solvers.

\section{Discussion}

\subsection{\added{Analytically guided architecture performance}}

\added{A key finding is the substantially improved prediction accuracy achieved by Model~3, in which the coefficient of consolidation $C_v$ is provided to the trunk net rather than the branch net (as in Models~1 and~2). This improvement is supported by the close correspondence between Model~3 and the analytical solution structure of the consolidation equation (\Cref{eq:terzaghi_analytical_solution}), given by}

\begin{equation}
u(z,t;C_v)
=
\sum_{a=0}^{\infty}
\underbrace{\frac{2u_0}{A}}_{b_a(u_0)}
\;
\underbrace{
\sin\!\left(\frac{Az}{H_{\mathrm{dr}}}\right)
\exp\!\left(-A^2 \frac{C_v t}{H_{\mathrm{dr}}^2}\right)
}_{\phi_a(z,t,C_v)}
\;\approx\;
\sum_{k=1}^{q}
b_k(u_0)\,t_k(z,t,C_v)
\end{equation}

\added{Where $b_a(u_0)$ depends only on the initial condition, and $\phi_a(z,t,C_v)$ are spatio-temporal basis functions parametrized by $C_v$. The right-hand side represents the corresponding DeepONet approximation, in which
$b_k$ and $t_k$ are learned by the branch and trunk networks, respectively. Notably, $C_v$ appears only in the exponential decay term through the dimensionless time factor $T_v=\frac{C_v t}{H_{\mathrm{dr}}^2}$, directly controlling the temporal evolution of the basis functions. Embedding $C_v$ in the trunk net therefore enables the learned basis functions $t_k(z,t,C_v)$ to adapt to different consolidation rates in a manner consistent with the analytical solution. This architectural alignment allows Model~3 to more effectively capture the influence of $C_v$ on the spatio-temporal evolution of the excess PWP field.}

From a geotechnical perspective, $C_v$ encapsulates the combined effects of soil permeability and compressibility, which are fundamental in predicting the overall consolidation timeline. An architecture that directly learns this influence on the temporal basis functions, like Model 3 does, is therefore more likely to align with the physical principles of the consolidation process. Moreover, Fourier feature embedding effectively improves the accuracy of Model 3 (i.e., Model 4) (\Cref{fig:modelsworst}), particularly in the early stages of consolidation. By mapping the low-dimensional inputs $(z, t, C_v)$ into a higher-dimensional space via sinusoidal functions, Fourier feature embedding enriches the input representation and enables the network to more readily approximate functions with high-frequency content or sharp spatio-temporal patterns. This architectural refinement accounts for Model 4’s superior accuracy, especially in challenging scenarios with steep excess PWP gradients. These observations can guide the implementation of future DeepONet architectures in geotechnical problems with similar PDE-driven systems governed by system coefficients with steep functional variations.

\subsection{Out-of-distribution response}

\begin{figure}[!htbp]
    \centering
    \includegraphics[width=0.65\textwidth]{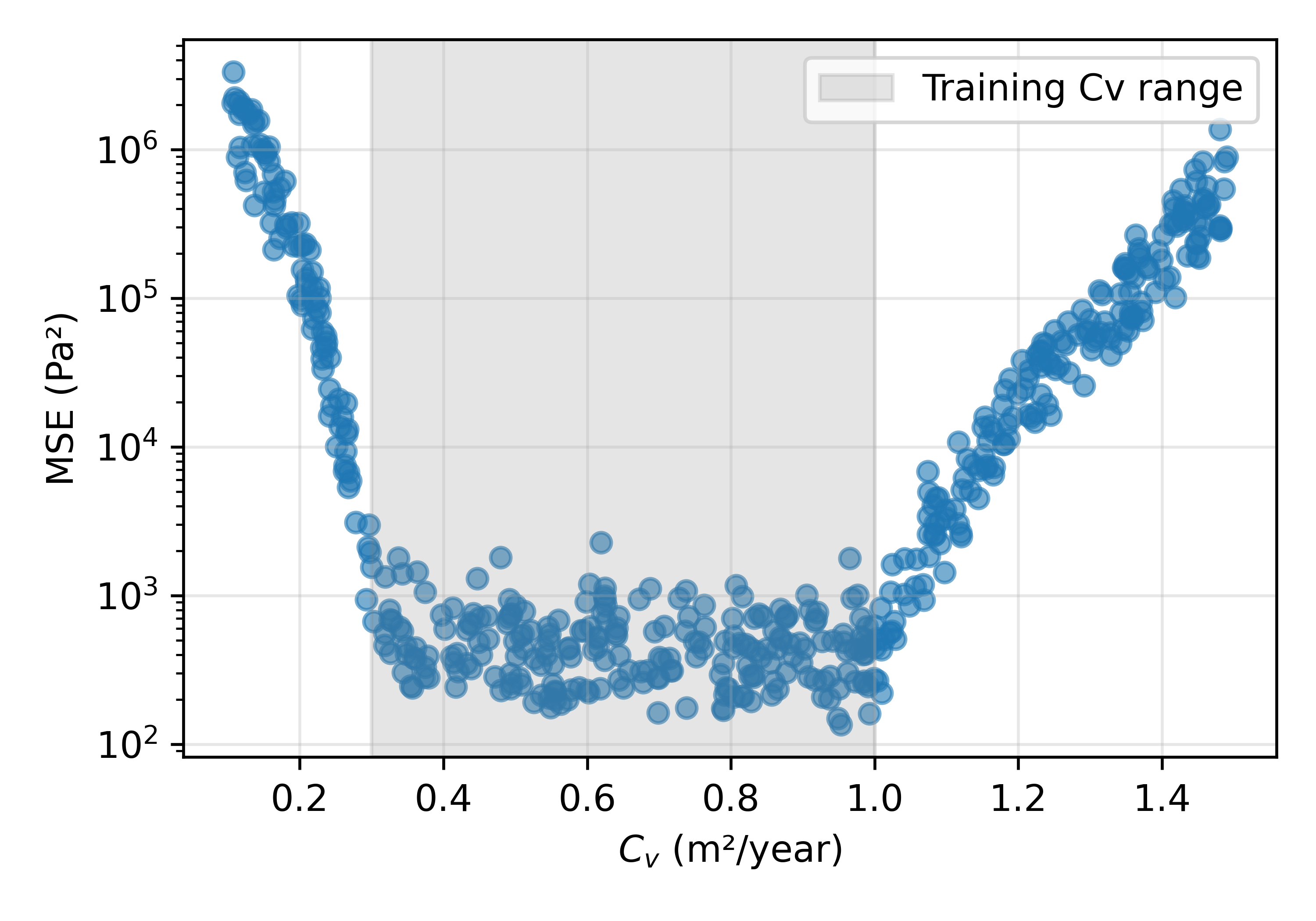}
    \caption{Mean squared error evaluated on $C_v$s with out-of-training-distribution values for Model 4.}
    \label{fig:mse_vs_cv_extended}
\end{figure}

We also evaluate the performance of the implemented DeepONet architecture on out-of-training distribution data focusing on 1D case. First, we assess the solution accuracy of Model 4 for values of $C_v$ outside the training range (\Cref{fig:mse_vs_cv_extended}). For this, we create an extra test data set with $C_v$ randomly sampled from 0.1 to 1.5 $m^2 / year$. Within the training range of $C_v$ (0.3 to 1.0 $m^2/year$, indicated by the shaded region), the model maintains low MSE, typically below $10^3 \ Pa^2$. However, as $C_v$ goes into the out-of-distribution region, the MSE shows an increase. This behavior is not unexpected (e.g., \citep{lu2021learning,ahmed2024deeponet_structure}) because DeepONet is an effective parameter-aware interpolator rather than an extrapolator. This implies that the successful deployment of a DeepONet-based surrogate model requires a careful selection of PDE parameters in the training dataset, which, in turn, requires relevant domain knowledge.

% Talk about Cv extrapolation

\begin{figure}[!htbp]
    \centering
    \includegraphics[width=0.7\textwidth]{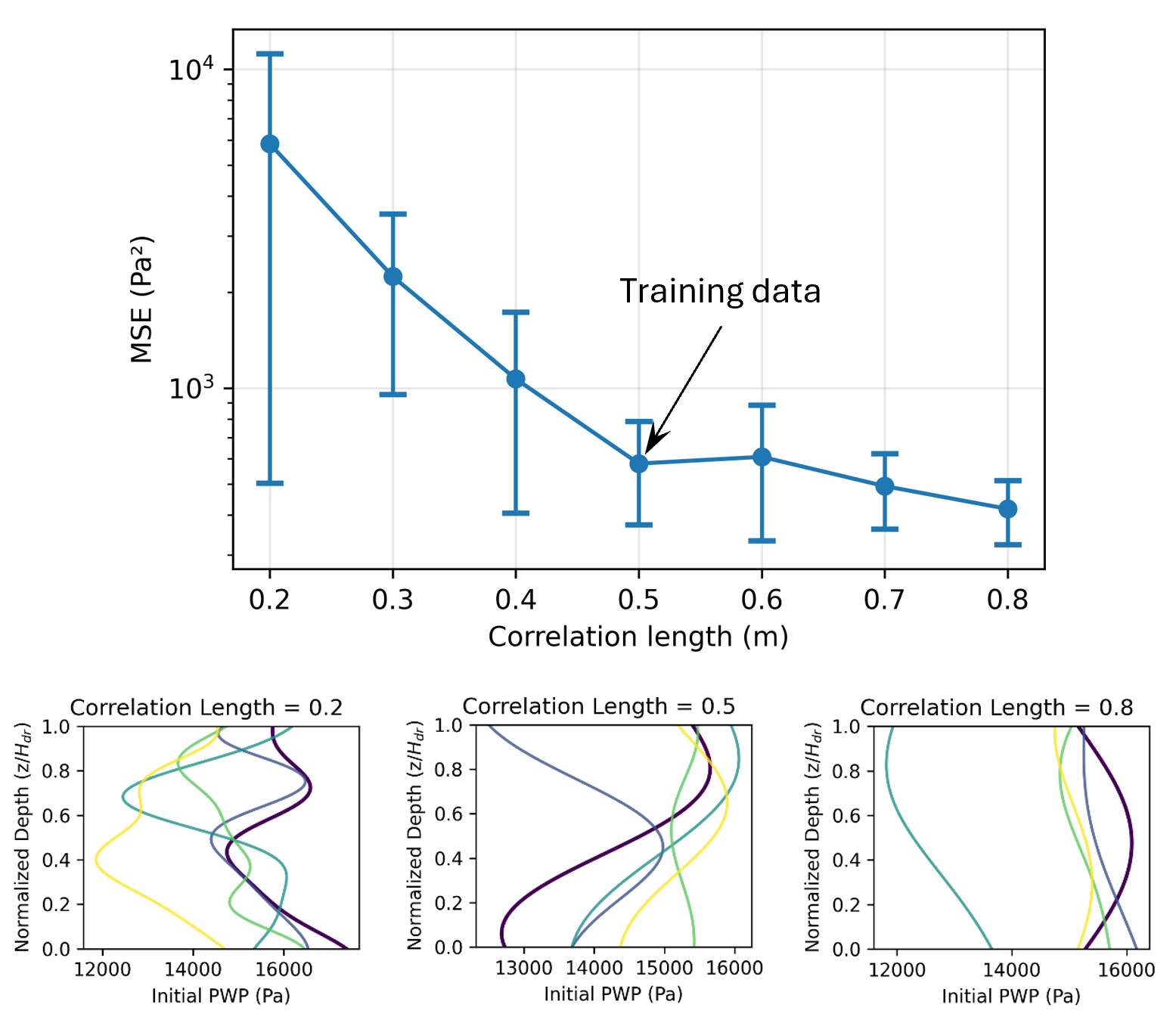}
    \caption{Mean squared error evaluated on different function spaces with out-of-training-distribution correlation lengths (0.2 to 0.8) for Model 4. Vertical bars indicate one standard deviation. The bottom plots show several realizations of Gaussian random fields for correlation lengths of 0.2, 0.5, and 0.8, illustrating how the input function varies with different correlation length values.}
    \label{fig:mse_vs_lengthscale}
\end{figure}

% Talk about function space extrapolation
We further evaluate the solution accuracy of Model 4 on function spaces characterized by correlation lengths ranging from 0.2 to 0.8, extending beyond the training correlation length of 0.5. For each correlation length, we construct an additional test dataset comprising 10 realizations of Gaussian random fields as input functions. The lower panel in \Cref{fig:mse_vs_lengthscale} shows examples of these functions at the correlation lengths 0.2, 0.5, and 0.8. Consistent with prior work on DeepONet extrapolation \citep{lu2021learning}, the prediction error increases when testing on functions with correlation lengths smaller than the 0.5 used in training. However, the prediction remains within a reasonable range; the mean MSE at a correlation length of 0.2 (approximately $6\times10^3 \ Pa^2$) is still lower than what we discussed in \Cref{fig:model3worst}, which exhibits an accurate solution except for the early times. Notably, the error does not increase for correlation lengths greater than 0.5, but decreases. This stability likely stems from the fact that GRFs with larger correlation lengths are smoother and more similar to the training function space, which spans 0.5 GRFs and uniform excess PWP functions. As a result, the model effectively interpolates between these GRFs and the constant functions. Along with \Cref{fig:mse_vs_cv_extended}, the result in \Cref{fig:mse_vs_lengthscale} highlights the importance of careful DeepONet training data design based on domain knowledge.

While this study primarily focuses on evaluating DeepONet's potential as an accurate solver for consolidation problems, through an architectural exploration, future work can address extrapolation limitations by incorporating physics-informed and adaptive training strategies, i.e., including domain expertise to ensure plausible ranges of $C_v$ and spatial pore pressure distributions are represented into the trained DeepONet for different geologies of interest.

\added{Alternatively, a physics-informed DeepONet \citep{wang2021learning_pideeponet}, in which the governing consolidation equation (\Cref{eq:terzaghi_pde}) is incorporated as a soft constraint in the loss function, could also enhance physical consistency and improve generalization outside the training range. In this setting, automatic differentiation is used to compute the residual of the PDE at collocation points in the spatio-temporal domain, and an additional physics-based loss term can be introduced to penalize violations of the PDE and boundary conditions, similar to the approaches from PINNs \citep{raissi2019pinn}. For the present consolidation problem, such a formulation could explicitly enforce diffusion-driven dissipation and pore pressure decay with their parameter dependence, potentially improving out-of-distribution errors. However, balancing data loss and physics-based constraints remains challenging, and improper weighting of them may lead to biased solutions or slow convergence \citep{faroughi2022sciml}.}

\added{In addition, active learning frameworks offer a promising strategy for adaptively enriching the training dataset in regions associated with high uncertainty or large prediction errors \citep{guan2022adaptive_active, tharwat2023survey_active}. For consolidation problems, this could be implemented by iteratively identifying ranges of $C_v$, correlation lengths, or early-time intervals where models exhibit elevated MSE, and selectively generating additional numerical simulations in these regimes. The newly sampled cases can then be incorporated into subsequent training rounds to improve coverage of challenging regions in the input and parameter spaces. Such an approach is expected to enhance generalization while minimizing the computational cost of data generation without full retraining. Nevertheless, active learning presents complexities in uncertainty estimation and sample selection, as its effectiveness is often linked to the reliability of the metrics used to gauge model confidence \citep{winovich2025active}.}

\subsection{Performance in the context of other operator-learning neural networks}

%%% \added: FNO brief intro------------------------------------------------
Besides DeepONet, there are also other operator-learning neural networks such as Fourier neural operator (FNO) \citep{li2020fourierNOs}, and transformer-based neural operators (TBNOs) \citep{hao2023transformer_no} that could be potentially considered for the consolidation problem. FNO learns operators using spectral convolutions: it maps input functions to Fourier space, applies learnable filters to selected frequency modes, and transforms back to the physical domain, enabling efficient learning of smooth global dependencies of the domain. A detailed description is provided in \cite{li2020fourierNOs}. TBNOs extend the attention mechanisms \citep{vaswani2017attention} of transformers, which were originally developed to learn sequential dependencies, to operator learning in space and time. \cite{hao2023transformer_no} provides additional details. Despite their increased learning capacity, TBNOs typically require more computational resources due to their use of complex neural network structures such as transformers \citep{hao2023transformer_no,li2022transformer_no}. In this study, we limit the additional exploration of alternative operator-learning neural networks to FNOs, which provide a well-established and computationally practical baseline comparable to DeepONet.

%%% \added: explain FNO structure------------------------------------------------
We construct the FNO on a $100 \times 100$ uniform grid spanning the normalized depth and time, $\frac{z}{H_{dr}} \in [0, 1]$ and $T_v=\frac{C_v t}{H_{dr}^2} \in [0, 2]$, the same range as DeepONet. Each grid point is represented by four input channels $C_v$, $z$, $t$, and $u_0$. The architecture comprises four spectral convolution layers with a hidden channel dimension of 8, followed by an output layer that produces $100 \times 100$ field representing the excess PWP $u(z,t)$. Thus, a single training sample consists of an input tensor of shape $100 \times 100 \times 4$ and a corresponding target solution tensor of shape $100 \times 100 \times 1$.

\begin{figure}[!htbp]
    \centering
    \includegraphics[width=1.0\textwidth]{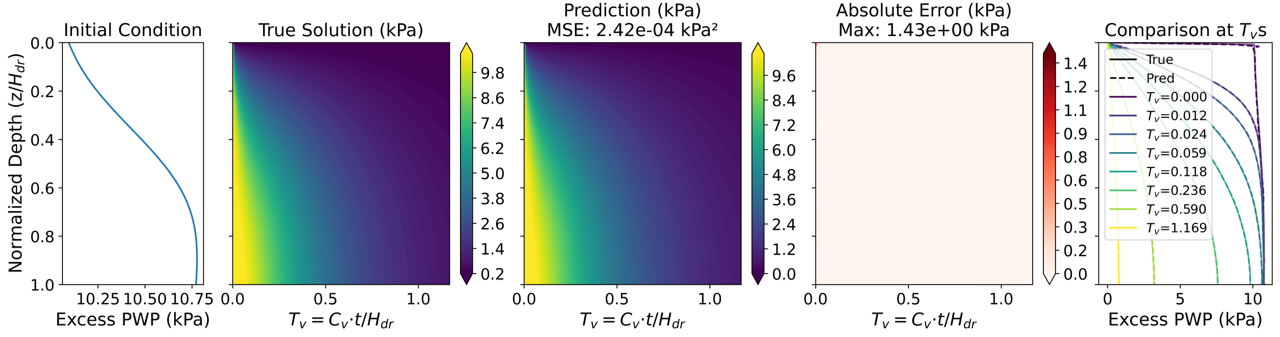}
    \caption{Worst case predictions from FNO with $C_v = 0.3009 \ m^2/year$. The plots share the y-axis.}
    \label{fig:fno_worst}
\end{figure}

The model is trained with the AdamW optimizer using a learning rate of 0.01, a batch size of 256, and 800 epochs. The training dataset contains 40,000 samples, and both the validation and test sets contain 5,000 samples. These training hyperparameters align with those used for Model 4. The resulting MSEs on the training, validation, and test sets are 116.49 $\text{Pa}^2$, 119.21 $\text{Pa}^2$, and 481.69 $\text{Pa}^2$ (with standard deviation of 153.63 $\text{Pa}^2$, 155.82 $\text{Pa}^2$, and 186.15 $\text{Pa}^2$), respectively. \Cref{fig:fno_worst} presents the worst-case prediction, like we showed in \Cref{fig:modelsworst}.

Although the FNO achieves lower training and validation MSEs than Model 4 (the enhanced DeepONet), its test MSE (481.69 $\text{Pa}^2$) does not show a meaningful difference from Model 4’s test MSE (508.96 $\text{Pa}^2$ \Cref{fig:mean_mse}). This indicates that, in our problem setting, while the FNO fits the training distribution more effectively, its improved representation performance does not translate into substantially better generalization to unseen functions. 

A contributing factor is the effective volume of supervised data each model receives. While both architectures are trained with 40,000 input functions, the supervision density differs due to their inherent architectural design: DeepONet observes only $10\times 10$ sparse solution samples per function, whereas FNO is trained on dense $100\times100$ regular solution fields spanning over the solution domain. Thus, FNO effectively receives far more labeled output data per function, which partly explains its lower prediction errors.

The training efforts also vary. Training FNO took approximately 165 minutes, over four times longer than Model 4 (around 40 minutes). This difference shows that the DeepONet can achieve comparable generalization accuracy with lower training costs for the consolidation problem being examined. Sensitivity to architecture and problem type might be a factor, so this is not a general statement for these operator learning frameworks. Such comparisons are beyond the scope of this study and can be conducted for a broader set of geotechnical problems in future efforts.

\added{DeepONet’s inference time and memory usage scale with the number of query points, whereas FNO’s computational cost primarily depends on the output grid resolution due to their inherent architectural differences. In our test setting, with a $50 \times 50$ output field, both DeepONet (Model~4) and FNO enable fast evaluation once trained, with inference costs that are negligible compared to classical numerical solvers. Specifically, Model~4 requires approximately $2.43\,\mathrm{s}$ to generate predictions for the entire test set (i.e., 500 solution fields), while FNO requires $4.12\,\mathrm{s}$. In terms of memory consumption, DeepONet uses about $120\,\mathrm{MB}$ of GPU memory, whereas FNO requires approximately $374\,\mathrm{MB}$. This difference is likely due to the relatively small size of Model~4, which allows it to outperform the global Fourier transforms employed by FNO on a $50 \times 50$ grid.}

Even though DeepONet and FNO performance were comparable for the consolidation problem, there are practical advantages for DeepONet over FNO. \added{FNO assumes data defined on uniform Cartesian grids in order to leverage Fast Fourier Transform (FFT)-based convolutions. Extensions to irregular geometries or non-uniform meshes typically require additional preprocessing, mapping, or architectural modifications \citep{li2023fourier_geom}. DeepONet, by contrast, does not impose grid regularity constraints and can naturally accommodate non-uniform meshes or irregular output sampling through its pointwise trunk inputs at queried locations. This allows DeepONet to be more flexible in irregular geometries \citep{lu2021learning,faroughi2022sciml}.} Additionally, DeepONet's flexible branch-trunk architecture accommodates diverse neural modules (e.g., MLPs, convolution neural networks \citep{mei2024cnn_don}, graph networks \citep{zhang2025graph_don}, and attention mechanisms \citep{chen2025attension_don}), facilitating hybrid designs and multimodal operator inputs. The training for DeepONet was also shorter for the configuration in this study. 
Finally, DeepONet is also supported by the theoretical foundation, universal approximation theorem for nonlinear operators in infinite-dimensional spaces \citep{chen1995universal,lu2021learning}.

\subsection{DeepOnet in the context of classical surrogate modeling}

\added{Classical surrogate modeling approaches, such as projection-based reduced-order models (ROMs) and polynomial chaos expansions (PCEs), are commonly used to approximate high-fidelity simulations, particularly for problems with low-dimensional parametric inputs or limited solution variability. However, their accuracy and computational efficiency often deteriorate as the dimensionality of the parameter space and the spatial complexity of the solution increase \citep{benner2015survey-rom}. Moreover, projection-based ROMs rely on a fixed spatial discretization, which limits their flexibility when transferring models across varying meshes during inference. In contrast, DeepONet is not restricted to a prescribed discretization, enabling greater adaptability to heterogeneous spatial representations \citep{dummer2026ronomreducedorderneuraloperator}.}

\added{To illustrate the performance of DeepONet in the context of classical ROMs, we compare DeepONet (Model~4) with a proper orthogonal decomposition (POD)-based ROM \citep{xiao2017parameterized,gao2024machine-rom} in a three-dimensional setting. The ROM is constructed by computing a truncated POD basis via singular value decomposition of full-field solution snapshots. Retaining 50 modes yields a set of orthonormal spatial basis functions, $\boldsymbol{\Phi} \in \mathbb{R}^{n \times 50}$\nomenclature{$\boldsymbol{\Phi}$}{Orthonormal basis for reduced order model}. An MLP with 256 hidden units maps the flattened initial excess pore water pressure $u_0$, the coefficient of consolidation $C_v$, and time $t$ to the corresponding POD coefficients $\boldsymbol{c} \in \mathbb{R}^{50}$\nomenclature{$\boldsymbol{c}$}{Proper orthogonal decomposition coefficients}. In this ROM formulation, the spatial basis $\boldsymbol{\Phi}$ is precomputed during training and remains fixed during inference, while only the coefficient mapping is learned. By contrast, DeepONet simultaneously learns both the basis functions (through the trunk network) and the associated coefficients (through the branch network), with the trunk network evaluated at each output location $(x,y,z,t)$. This operator-learning formulation enables DeepONet to represent more flexible and adaptive solution manifolds.}

\begin{figure}[!htbp]
    \centering
    \includegraphics[width=1.0\textwidth]{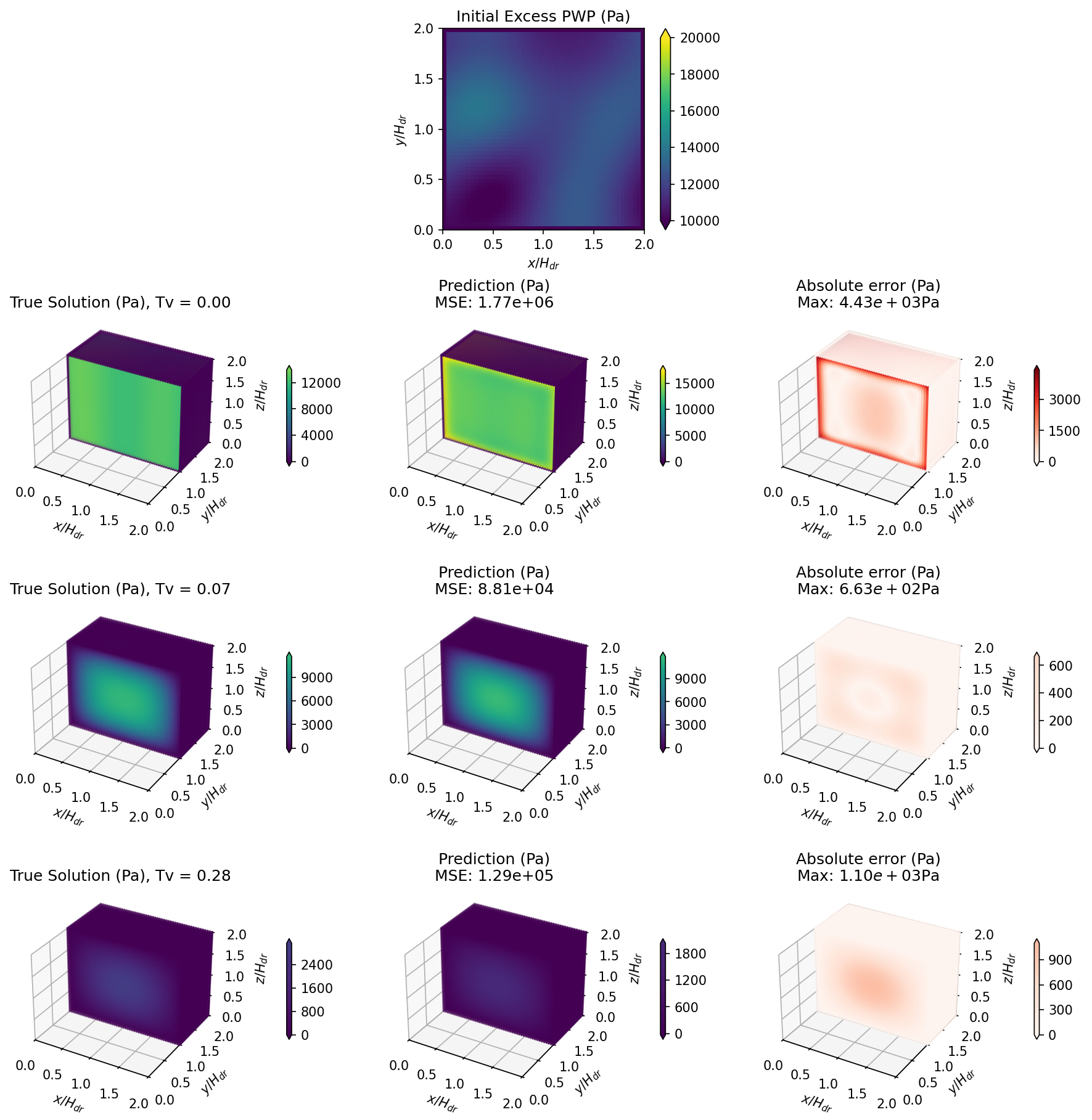}
    \caption{Excess PWP dissipation evaluated by the ROM on test data at $C_v=0.35 \ m^2/year$ with spatially varying initial excess PWP. The subsequent rows show the solution field from the finite difference method, prediction from ROM, and the absolute error, at the specified $T_v$}
    \label{fig:ROM-result}
\end{figure}

\added{We evaluate the ROM on 10 uniformly sampled test cases under the same conditions used for DeepONet. The ROM attains a mean field mean-squared error (MSE) of $1.528\times10^{6},\mathrm{Pa}^2$, with a mean prediction time of 0.105~s per full three-dimensional field. A representative prediction is shown in \Cref{fig:ROM-result}. While the computational cost is comparable to that of DeepONet (\Cref{table:comp_time_3d}), the ROM consistently exhibits larger prediction errors (\Cref{fig:three_d_result}), particularly at early times and in regions where the initial excess pore water pressure exhibits strong spatial gradients.}

\added{Several caveats apply to this comparison. Data equivalence is defined in terms of total storage requirements: the ROM is trained using 250 full-field input–output pairs, which occupy approximately the same storage as the DeepONet training dataset. However, the two approaches rely on fundamentally different data representations. The ROM requires dense full-field snapshots for offline basis construction, whereas DeepONet is trained using sparsely sampled pointwise evaluations, complicating a strictly equitable data comparison. Furthermore, the effective model capacities are governed by distinct and non-commensurable hyperparameters, including the number of retained POD modes and the MLP architecture for the ROM, versus the branch-trunk network widths and Fourier feature settings for DeepONet. A systematic exploration of these factors is beyond the scope of this study and represents an important direction for future work.}

\section{Conclusion}
This study demonstrates the performance of the DeepONet for learning solution operators in consolidation problems. Focusing first on the 1D case, we systematically investigate different DeepONet architectures. Our results reveal that feeding $C_v$ into the trunk net (Model 3), rather than the more typical branch net (models 1 and 2), substantially improves prediction accuracy by 6 to 12 times lower test MSE. The improved accuracy is supported by the analytical form of the PDE, where $C_v$ modulates time-dependent behavior in the basis function. In addition, we further enhance the DeepONet architecture in Model 3 using Fourier Feature Embedding (Model 4). This further reduces the error by 1.5 times that of Model 3, achieving the highest accuracy among all models. The architecture in Model 4 also effectively mitigates the high errors shown in highly varying, early-time excess PWP behaviors in the other model, due to its enhanced capacity to represent sharp spatio-temporal features from Fourier feature embedding. These findings offer guidance for designing DeepONet architectures for operator learning in PDE-driven systems governed by physical coefficients. Finally, we extend Model 4 to the 3D consolidation problem and demonstrate its use in an uncertainty quantification example.

The trained DeepONet models achieve a significant computational speedup compared to traditional numerical solvers used to generate training data and benchmarks. In the 3D problem, DeepONet generates full 3D fields over time in less than a second, showing about three orders of magnitude faster computation speed than the high-fidelity finite difference-based solver. Although the absolute time differences are on the order of milliseconds for the 1D consolidation case, the relative speedup,ranging from 1.5 to 100 times faster, is still meaningful. Leveraging this efficiency, we show a conceptual demonstration of DeepONet’s potential to accelerate uncertainty quantification in 3D consolidation. This computational advantage could be particularly valuable in large-scale or high-dimensional geotechnical simulations, where conventional solvers incur high costs due to time-stepping and meshing overhead.

Our results also show that the trained DeepONet model performance diminishes when extrapolating to out-of-distribution values of $C_v$ and the function spaces for the initial excess PWP condition. This suggests that the deployment of a DeepONet-based surrogate model requires careful PDE parameter selection of a training dataset, which, in turn, requires relevant geotechnical expert knowledge. If the problem can be described with a full PDE form, physics-informed loss terms can be employed to mitigate these output distribution errors (e.g., \cite{wang2021learning_pideeponet}). Alternatively, active learning frameworks (e.g., \cite{guan2022adaptive_active, tharwat2023survey_active}) could also be considered.

Beyond the highlighted technical contributions, this work introduces DeepONet to the geotechnical engineering community as an effective surrogate modeling framework. Its ability to generalize across varying PDE parameters and initial conditions without retraining makes it especially attractive for real-time prediction, uncertainty quantification, design optimization, and inverse problems, where traditional numerical solvers or even physics-informed neural networks may fall short due to computational or formulation constraints.

\added{
In terms of engineering applications, we believe that the DeepONet model implemented in this study has the potential to be integrated with modern sensing technologies to enable pore-pressure monitoring and subsurface characterization, thereby supporting a real-time monitoring framework for consolidation-related problems (e.g., embankment consolidation, wick-drain performance). Although the example presented in the manuscript is based on synthetic data, it demonstrates DeepONet's efficiency. Future studies that collect dense pore-pressure data and monitor the evolution of consolidation over time would be ideal for further advancing and validating the framework envisioned in this study.
}

\added{
Additionally, while this study uses a constant $C_v$ as trunk input, spatial heterogeneity could be addressed in future efforts by treating the soil property field $C_v(x,y,z)$ as an additional input function to the Branch network. Finally, a potential pathway for experimental integration could involve transfer learning, where a model pre-trained on large-scale synthetic data is fine-tuned using sparse, high-fidelity field measurements, providing a multi-fidelity DeepONet \citep{xu2024multi}.
}

\section*{Acknowledgments}
This material is based on work supported by the National Science Foundation (NSF) under Grant No. CMMI 2145092. Any opinions, findings, conclusions, or recommendations expressed in this material are those of the author(s) and do not necessarily reflect the views of the NSF. This work is also supported by the InnoCORE program of the Ministry of Science and ICT (N10250154).

\nomenclature{$N_x, N_y, N_z$}{The number of discretization nodes in the x, y, and z directions for finite difference}
\nomenclature{$\Delta x, \Delta y, \Delta z$}{Grid spacings for spatial discretization for finite difference}
\nomenclature{$\Delta t$}{The time step size}

\printnomenclature

\appendix
\section{Numerical methods to solve \Cref{eq:discretized_odes}}
\label{sec:appendix}

The simplest BDF method (first-order backward Euler differentiation) can be expressed in matrix form as:

\begin{equation}
    (\mathbf{I} - \Delta t \mathbf{A})\mathbf{u}^{t+1} = \mathbf{u}^{t}
    \label{eq:bdf1}
\end{equation}
where $ \mathbf{I} $ is the identity matrix, and $ \mathbf{u}^{t} $ and $ \mathbf{u}^{t+1} $ represent vectors of excess PWPs at the current and next time steps, respectively. The matrix $ \mathbf{A} $ represents the central difference approximation for the second derivative, which arises from the spatial discretization. It has a tridiagonal matrix form:

\begin{equation}
    \mathbf{A} = \frac{C_v}{\Delta z^2}
    \begin{bmatrix}
        -2 & 1 & 0 & \dots & 0\\[5pt]
        1 & -2 & 1 & \dots & 0\\[5pt]
        0 & 1 & -2 & \dots & 0\\[5pt]
        \vdots & \vdots & \vdots & \ddots & 1\\[5pt]
        0 & 0 & 0 & 1 & -2
    \end{bmatrix}
    \label{eq:matrixA}
\end{equation}

At each time step, the solver implicitly solves \Cref{eq:bdf1} for the unknown PWP distribution $ \mathbf{u}^{t+1} $. In practice, the solver adaptively selects an integration order (up to fifth-order BDF) according to prescribed error tolerances. The general formulation of higher-order BDF methods is given by:

\begin{equation}
    \sum_{j=0}^{k}\alpha_j\mathbf{u}^{t+1-j} = \Delta t\,\mathbf{A}\mathbf{u}^{t+1}
    \label{eq:bdf_general}
\end{equation}

where $ \alpha_j $ are method-specific coefficients determined by the selected order $ k $, which denotes the number of previous time steps used in the integration scheme. SciPy uses adaptive $k$ from 1 to 5 for the optimized performance based on estimated local error and step size control.

Unlike implicit schemes, the Runge-Kutta (RK) integration method does not require solving a linear system at each step. The general form of the explicit integration step is given as:

\begin{equation}
\mathbf{u}^{t+1} = \mathbf{u}^{t} + \Delta t\,f(t_n,\mathbf{u}^{t}),
\end{equation}

where $ f(\mathbf{u})=\mathbf{A}\mathbf{u} $ represents the spatially discretized PDE right-hand side. 

For instance, the classical fourth-order Runge-Kutta (RK4) method computes the next step explicitly as follows:

\begin{equation}
\begin{aligned}
k_1 &= f(t_n, \mathbf{u}^{t}), \\
k_2 &= f\left(t_n+\frac{\Delta t}{2}, \mathbf{u}^{t} + \frac{\Delta t}{2}k_1\right), \\
k_3 &= f\left(t_n+\frac{\Delta t}{2}, \mathbf{u}^{t} + \frac{\Delta t}{2}k_2\right), \\
k_4 &= f(t_n+\Delta t, \mathbf{u}^{t} + \Delta t\,k_3), \\
\mathbf{u}^{t+1} &= \mathbf{u}^{t} + \frac{\Delta t}{6}\left(k_1 + 2k_2 + 2k_3 + k_4\right)
\end{aligned}
\end{equation}

Explicit RK methods are simpler to implement and computationally inexpensive per step, but they typically require smaller time steps compared to implicit methods. SciPy’s \texttt{solve\_ivp} adaptively selects the time step size for improved efficiency. 



\bibliographystyle{elsarticle-harv} 
\bibliography{main}

%% TeX file.s 




% \end{thebibliography}
\clearpage

\end{document}